\pdfoutput=1

\documentclass[11pt]{article}

\usepackage[final]{acl}

\usepackage{times}
\usepackage{latexsym}
\usepackage{graphicx}
\usepackage{tabularx}
\usepackage{hyperref}
\usepackage{multicol}
\usepackage{multirow}
\usepackage{enumitem}
\usepackage{lscape}

\usepackage{booktabs}

\usepackage[T1]{fontenc}

\usepackage[utf8]{inputenc}

\usepackage{microtype}

\usepackage{inconsolata}
\usepackage{amsmath}

\usepackage{float}
\usepackage{soul}

\title{Free-text Rationale Generation under Readability Level Control}

\newcommand{\affilsup}[1]{\rlap{\textsuperscript{\normalfont#1}}}

\author{
    Yi-Sheng Hsu\affilsup{1,2,5}
    \qquad 
    Nils Feldhus\affilsup{1,3,4}
    \qquad 
    Sherzod Hakimov\affilsup{2}
    \\
    $^1$German Research Center for Artificial Intelligence (DFKI)
    \\
    $^2$Computational Linguistics, Department of Linguistics, Universit\"at Potsdam
    \\
    $^3$Quality and Usability Lab, Technische Universit\"at Berlin
    \\
    $^4$BIFOLD – Berlin Institute for the Foundations of Learning and Data
    \\
    $^5$Computer Science Institute, Hochschule Ruhr West
    \\
    {\small \texttt{yi-sheng.hsu@hs-ruhrwest.de} \hspace{5em} \texttt{feldhus@tu-berlin.de}}
}

\begin{document}
\maketitle

\begin{abstract}
    Free-text rationales justify model decisions in natural language and thus become likable and accessible among approaches to explanation across many tasks.
    However, their effectiveness can be hindered by misinterpretation and hallucination.
    As a perturbation test, we investigate how large language models (LLMs) perform rationale generation under the effects of readability level control, i.e., being prompted for an explanation targeting a specific expertise level, such as sixth grade or college.
    We find that explanations are adaptable to such instruction, though the observed distinction between readability levels does not fully match the defined complexity scores according to traditional readability metrics.
    Furthermore, the generated rationales tend to feature medium level complexity, which correlates with the measured quality using automatic metrics.
    Finally, our human annotators confirm a generally satisfactory impression on rationales at all readability levels, with high-school-level readability being most commonly perceived and favored.\footnote{\textbf{Disclaimer:} The article contains offensive or hateful materials, which is inevitable in the nature of the work.}

\end{abstract}

\section{Introduction}

Over the past few years, the rapid development of machine learning methods has drawn considerable attention to the research field of explainable artificial intelligence (XAI). While conventional approaches focused more on local or global analyses of rules and features \cite{DBLP:journals/corr/abs-1804-06620, DBLP:journals/tetci/ZhangTLT21}, the recent development of LLMs introduced more dynamic methodologies along with their enhanced capability of natural language generation (NLG). The self-explanation potentials of LLMs have been explored in a variety of approaches, such as examining free-text rationales \cite{DBLP:conf/emnlp/WiegreffeMS21} or combining LLM output with saliency maps \cite{DBLP:journals/corr/abs-2310-11207}.

Although natural language explanation (NLE) established itself to be among the most common approaches to justify LLM predictions \cite{zhu-2024-llm-explanation-tutorial}, free-text rationales were found to potentially misalign with the predictions and thereby mislead human readers, for whom such misalignment seems hardly perceivable \cite{DBLP:conf/nips/YeD22}. Furthermore, it remains unexplored whether free-text rationales represent a model's decision making, or if they are generated just like any other NLG output regarding faithfulness. In light of this, we aim to examine whether free-text rationales can also be controlled through perturbation as demonstrated on NLG tasks \cite{dathathri-2020-plug-and-play, imperial-madabushi-2023-uniform}. If more dispersed text complexity could be observed in the rationales, it would indicate a higher resemblance between rationales and common NLG output, as we assume the LLMs to undergo a consistent decision making process on the same instance even under different instructions.

\begin{figure*}
    \centering
    \includegraphics[width=\textwidth, keepaspectratio]{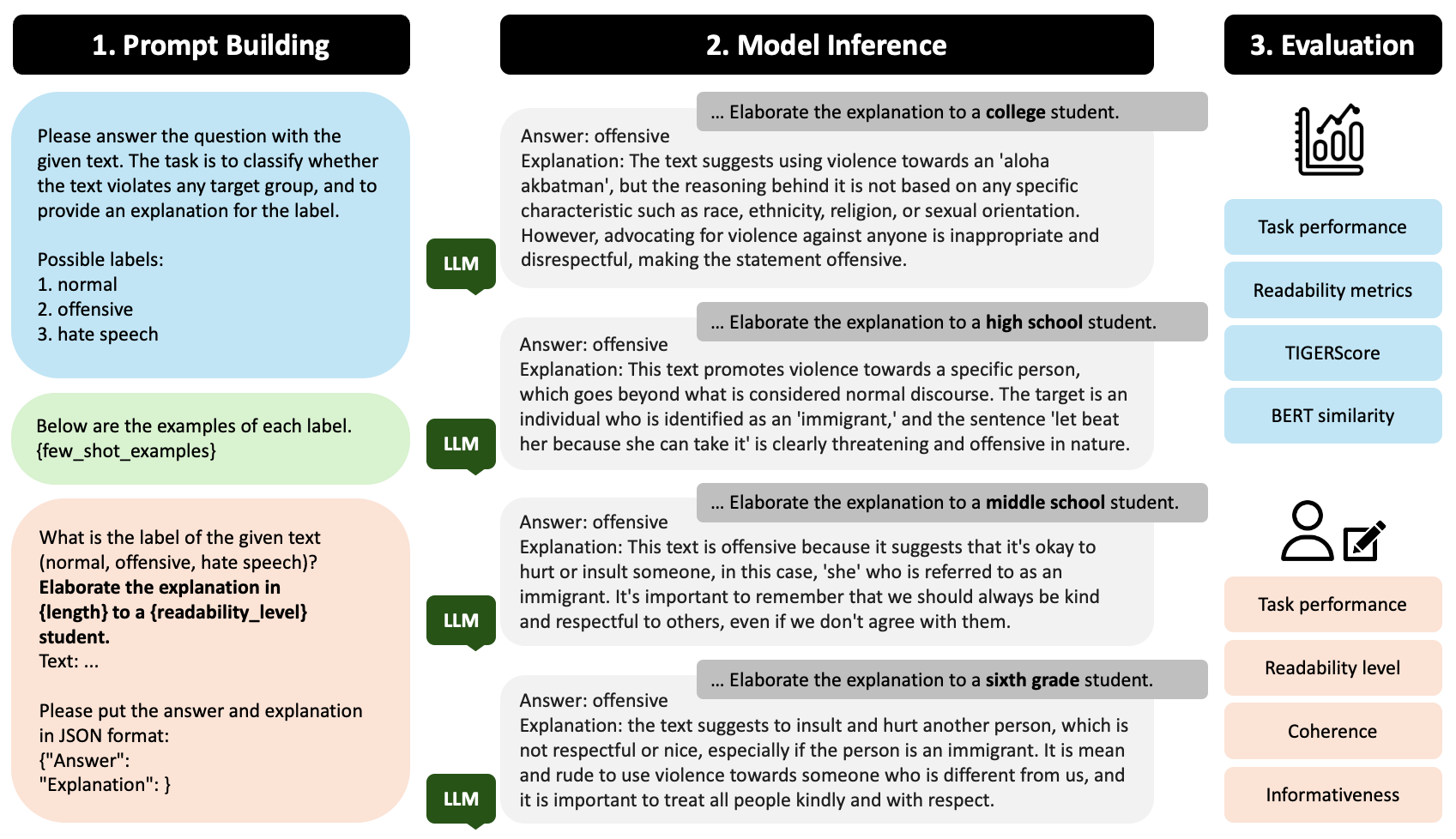}
    \caption{The experiment workflow of the current study. The demonstrated example comes from the \textsf{HateXplain} dataset. Generated responses are evaluated by both automatic metrics and human annotations.}
    \label{fig:workflow}
\end{figure*}




Targeting free-text rationales, we control text complexity with descriptive readability levels and evaluate the generated rationales under various frameworks to investigate what effects additional instructions or constraints may bring forward to the NLE task (Figure~\ref{fig:workflow}). Although the impact of readability \cite{DBLP:conf/acl/Stajner21} has rarely been addressed for NLEs, establishing such a connection could benefit model explainability, which ultimately aims at perception \cite{DBLP:conf/iui/EhsanTCHR19} and utility \cite{DBLP:conf/acl/JoshiLRCTNW0023} of diverse human recipients.

Our study makes the following contributions: First, we explore LLM output in both prediction and free-text rationalization under the influence of readability level control. Second, we apply objective metrics to evaluate the rationales and measure their quality across text complexity. Finally, we test how human perceive the complexity and quality of the rationales across different readability levels.\footnote{\textls[-60]{\url{https://github.com/doyouwantsometea/nle_readability}}}

\section{Background}

\paragraph{Text complexity}

The notion of text complexity was brought forward in early studies to measure how readers of various education levels comprehend a given text~\cite{kincaid1975derivation}. Prior to recent developments of NLP, text complexity was approximated through metrics including Flesch Reading Ease (FRE, \citealp{kincaid1975derivation}), Gunning fox index (GFI, \citealp{gunning1952technique}), and Coleman-Liau index (CLI, \citealp{coleman1975computer}) (Appendix~\ref{appendix:metric}). These approaches quantify readability through formulas considering factors like sentence length, word counts, and syllable counts.

As the most common readability metric, FRE was often mapped to descriptions that bridge between numeric scores and educational levels \cite{farajidizaji-2024-possible-modify}. 
\citet{ribeiro-2023-generating-summaries} applied readability level control to text summarization through instruction-prompting. In their study, descriptive categories were prompted for assigning desired text complexity to LLM output.

\paragraph{NLE metrics}

Although the assessment of explainable models lacks a unified standard, mainstream approaches employ either objective or human-in-the-loop evaluation \cite{DBLP:journals/inffus/ViloneL21}. 
Objective metric scores include LAS \cite{DBLP:conf/emnlp/HaseZXB20}, REV \cite{DBLP:conf/acl/ChenB0J0S23}, and RORA \cite{DBLP:conf/acl/JiangLCKDL24}. 
Their training processes highly rely on a particular data structure, which does not generalize to tasks relevant to readability. Furthermore, while most studies on NLE intuitively presume model-generated rationales to bridge between model input and output, it remains unclear whether the provided reasoning faithfully represents its internal process for output generation; in other words, free-text rationales could be only reflecting what the model has learned from its training data \cite{atanasova-etal-2023-faithfulness}.


\begin{table}[t]
    \centering
    \renewcommand*{\arraystretch}{0.9}
    \resizebox{\columnwidth}{!}{
    \begin{tabular}{r|cccc}
        \textbf{FRE} & >80 & 60-80 & 40-60 & <40 \\
        \midrule
        \textbf{Readability} & sixth& middle & high & \multirow{2}{*}{college} \\
        \textbf{Level} & grade & school & school \\
    \end{tabular}
    }
    \caption{The mapping between FRE scores and readability levels adapted from \citet{ribeiro-2023-generating-summaries}.}
    \label{tab:FRE_level}
    \vspace*{-1em}
\end{table}

\section{Method}
\paragraph{Readability level control}
As demonstrated in Figure~\ref{fig:workflow}, in step~1, we incorporate instruction-prompting into the prompt building. The prompts consist of three sections: task description, few-shot in-context samples, and instruction for the test instance. After task description and samples, we add a statement aiming for the rationale: \textit{Elaborate the explanation in \{length\}\footnote{Throughout the experiments, we set this to a fixed value of \texttt{``three sentences''}.} to a \{readability\_level\} student.} 
Then we iterate through the data instances and readability levels in separate sessions. 
We adapt the framework of \citet{ribeiro-2023-generating-summaries} to four readability levels based on FRE score ranges (Table~\ref{tab:FRE_level})
and explore a range of desired FRE scores among \{30, 50, 70, 90\}, which are respectively phrased in the prompts as readability levels \{college, high school, middle school, sixth grade\}.

\paragraph{Evaluating free-text rationales}
In light of the problematic adaption to readability-related tasks and major issues in reproducibility of the aforementioned NLE evaluation metrics, we exploit the overlap between NLE and NLG, we adopt TIGERScore \cite{jiang-2024-tigerscore}, an NLG metric that is widely applicable to most tasks, for evaluating the generated free-text rationales (\S \ref{sec:evaluation}). Applying fine-tuned \texttt{Llama-2} \cite{DBLP:journals/corr/abs-2307-09288}, the metric was proposed to require little reference but instead rely on error analysis over prompted contexts to identify and grade mistakes in unstructured text. Nevertheless, the approach could sometimes suffer from hallucination (or confabulation), similar to the common LLM-based methodologies.

\begin{figure*}
    \centering
    \resizebox{\textwidth}{!}{%
        \includegraphics{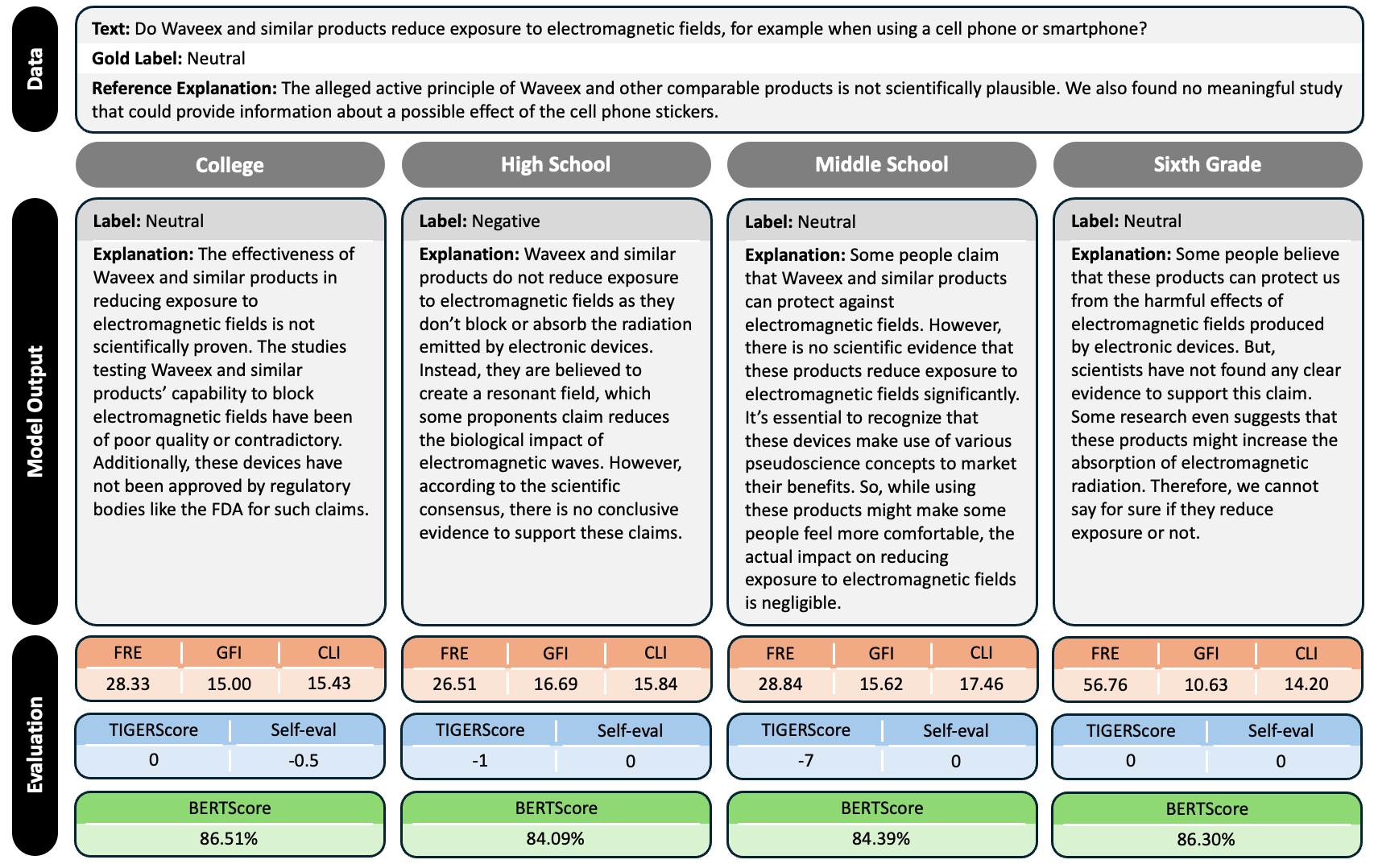}}
    \caption{An example of model predictions and rationales generated by \texttt{Mistral-0.2} on \textsf{HealthFC} along with the evaluation results. Self-eval refers to TIGERScore rated by \texttt{Mistral-0.2}.}
    \label{fig:output}
\end{figure*}

\section{Experiments}

\subsection{Rationale generation}
\label{sec:experiments:rationale_generation}

\paragraph{Datasets} We conduct readability-controlled rationale generation on three NLP tasks: fact-checking, hate speech detection, and natural language inference (NLI), adopting the datasets featuring explanatory annotations. For fact-checking, \textsf{HealthFC} \cite{DBLP:conf/coling/VladikaSM24} includes 750 claims for fact-checking under the medical domain, with excerpts of human-written explanations provided along with the verification labels.
For hate speech detection, two datasets are applied: (1) \textsf{HateXplain} \cite{mathew-2021-hatexplain}, which consists of 20k Tweets with human-highlighted keywords that contribute the most to the labels. 
(2) Contextual Abuse Dataset (\textsf{CAD}, \citealp{DBLP:conf/naacl/VidgenNMRT21}), which contains 25k entries with six unique labels elaborating the context under which hatred is expressed. 
Lastly, \textsf{SpanEx} \cite{choudhury-etal-2023-explaining} is an NLI dataset that includes annotations on word-level semantic relations  (Appendix~\ref{appendix:dataset}).

\paragraph{Models} We select four recent open-weight LLMs from three different families: \texttt{Mistral-0.2 7B} \cite{DBLP:journals/corr/abs-2310-06825}, \texttt{Mixtral-0.1 8x7B} \cite{DBLP:journals/corr/abs-2401-04088}\footnote{Owing to the larger size of \texttt{Mixtral-v0.1 8x7B}, we adopt a \texttt{bitsandbytes} 4-bit quantized version (\url{https://hf.co/ybelkada/Mixtral-8x7B-Instruct-v0.1-bnb-4bit}) to reduce memory consumption.}, \texttt{OpenChat-3.5 7B} \cite{DBLP:journals/corr/abs-2309-11235}, and \texttt{Llama-3 8B} 
\cite{DBLP:journals/corr/abs-2407-21783}. 
All the models are instruction-tuned variants downloaded from Hugging Face, using the default generation settings, running on NVIDIA A100 GPU.

\begin{figure*}[t]
    \centering
    \resizebox{.95\textwidth}{!}{%
        \includegraphics[width=5cm]{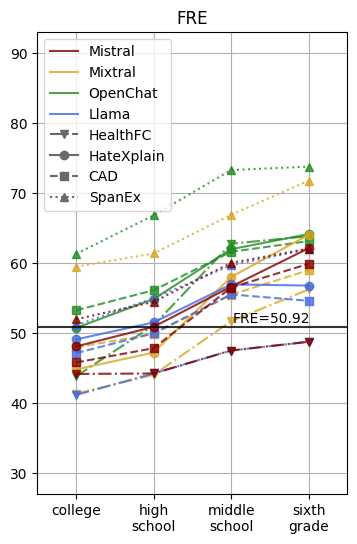}
        \includegraphics[width=5cm]{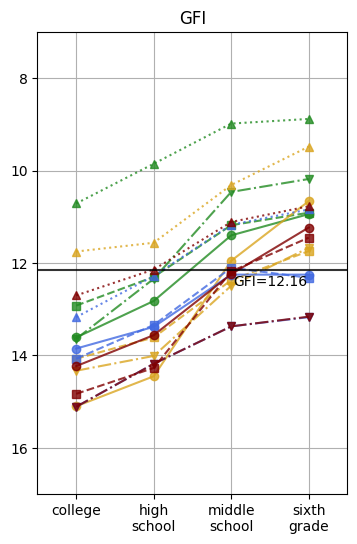}
        \includegraphics[width=5cm]{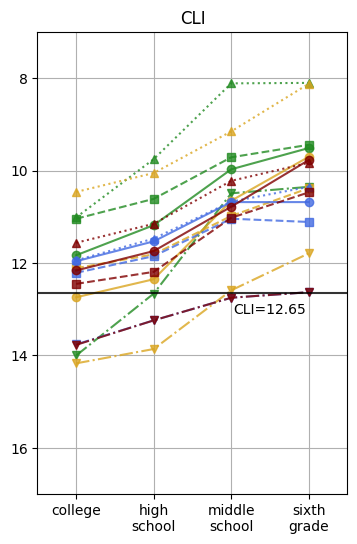}
    }
    \caption{The readability scores of model-generated rationales. Higher FRE score indicates lower text complexity, while GFI and CLI scores are in reverse. The black lines denote the readability scores of the reference rationales from \textsf{HealthFC}, which are provided in natural language instead of annotations (Appendix~\ref{appendix:dataset}).
    }
    \label{fig:readability_results}
\end{figure*}

\begin{figure*}
    \centering
    \resizebox{.95\textwidth}{!}{%
        \includegraphics[width=3.8cm]{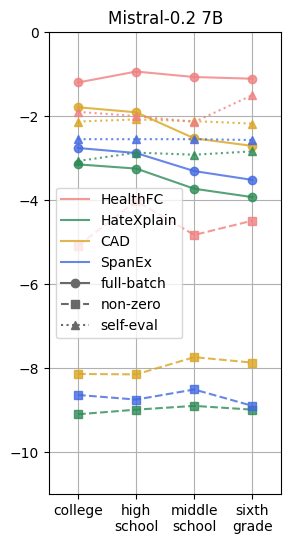}
        \includegraphics[width=3.8cm]{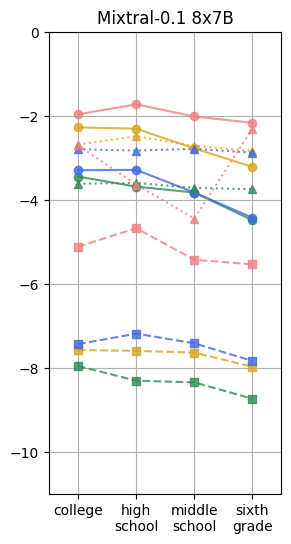}
        \includegraphics[width=3.8cm]{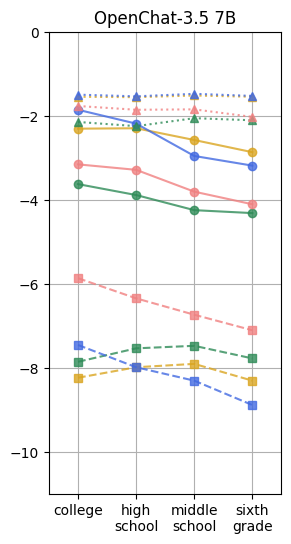}
        \includegraphics[width=3.8cm]{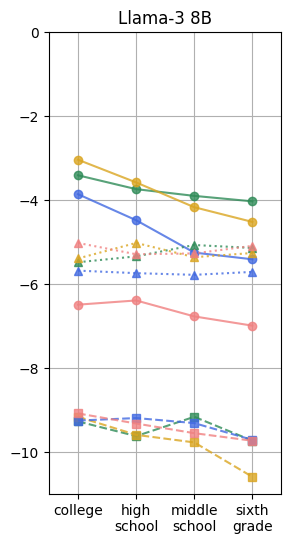}
    }
    \caption{TIGERScore evaluation results by model. Full-batch score reports the average of all data points, while the other two scores are divided by the amount of instances scoring below 0. The results of \texttt{Mistral-0.2} and \texttt{Mixtral-0.1} on \textsf{CAD} and \textsf{HealthFC} may induce more biases owing to the higher proportion of removed instances.
    }
    \label{fig:tigerscore_results}
\end{figure*}

\subsection{Evaluation}
\label{sec:evaluation}

\paragraph{Task accuracy} We use accuracy scores to assess the alignment between the model predictions and the gold labels processed from the datasets. In \textsf{HateXplain} \cite{mathew-2021-hatexplain}, since different annotators could label the same instance differently, we adopt the most frequent one as the gold label. Similarly, in \textsf{CAD} \cite{DBLP:conf/naacl/VidgenNMRT21}, we disregard the subcategories under ``offensive'' label to reduce complexity, simplifying the task into binary classification and leaving the subcategories as the source of building reference rationales.

\paragraph{Readability metrics} We choose three conventional readability metrics: FRE \cite{kincaid1975derivation}, GFI \cite{gunning1952technique}, and CLI \cite{coleman1975computer} to approximate the complexity of the rationales. While a higher FRE score indicates more readable text, higher GFI and CLI scores imply higher text complexity (Appendix \ref{appendix:metric}).

\begin{table}
    \centering
    \renewcommand*{\arraystretch}{}
    \resizebox{\columnwidth}{!}{%
    \begin{tabular}{ll|cccc|c}
        & \textbf{Readability} & 30 & 50 & 70 & 90 & Avg. \\
        \hline
        \parbox[t]{2mm}{\multirow{4}{*}{\rotatebox[origin=c]{90}{\small{\textbf{\textsf{HealthFC}}}}}} & \texttt{Mistral-0.2} & 52.8 & 52.8 & 53.8 & 50.2 & 52.4 \\
        & \texttt{Mixtral-0.1} & 54.7 & \textbf{56.4} & 55.0 & 55.9 & 55.5 \\
        & \texttt{OpenChat-3.5} & 51.6 & 53.0 & 52.8 & 51.8 & 52.3 \\
        & \texttt{Llama-3} & 27.9 & 30.9 & 30.0 & 27.8 & 29.2 \\
        \hline
        \parbox[t]{2mm}{\multirow{4}{*}{\rotatebox[origin=c]{90}{\small{\textbf{\textsf{HateXplain}}}}}} & \texttt{Mistral-0.2} & 49.4 & 49.3 & 52.6 & 52.0 & 50.8 \\
        & \texttt{Mixtral-0.1} & 46.1 & 48.4 & 47.2 & 47.5 & 47.3 \\
        & \texttt{OpenChat-3.5} & 51.7 & 51.5 & \textbf{53.0} & 50.5 & 51.7 \\
        & \texttt{Llama-3} & 50.7 & 51.4 & 50.5 & 50.3 & 50.7 \\
        \hline
        \parbox[t]{2mm}{\multirow{4}{*}{\rotatebox[origin=c]{90}{\small{\textbf{\textsf{CAD}}}}}} & \texttt{Mistral-0.2} & \textbf{82.3}* & 82.0 & 79.5 & 77.6 & 80.4 \\
        & \texttt{Mixtral-0.1} & 65.8* & 64.8 & 63.6 & 61.8 & 64.0 \\
        & \texttt{OpenChat-3.5} & 77.3 & 78.1 & 77.8 & 77.2 & 77.6 \\
        & \texttt{Llama-3} & 60.6* & 58.8 & 58.0 & 55.6 & 58.3 \\
        \hline
        \parbox[t]{2mm}{\multirow{4}{*}{\rotatebox[origin=c]{90}{\small{\textbf{\textsf{SpanEx}}}}}} & \texttt{Mistral-0.2} & 34.9 & 35.5 & 36.6 & 37.2 & 36.1 \\
        & \texttt{Mixtral-0.1} & 58.4 & 55.8 & 55.2 & 58.1 & 56.9 \\
        & \texttt{OpenChat-3.5} & 84.0 & 84.3 & 83.8 & \textbf{84.8}* & 84.2 \\
        & \texttt{Llama-3} & 41.8 & 41.7 & 42.0 & 41.1 & 41.7
    \end{tabular}
    }
    \caption{Task accuracy scores (\%) after removal of inappropriate answers. The highest score(s) achieved per model are starred, and best accuracy per task are highlighted in bold. Readability of 30, 50, 70, and 90 respectively refers to the desired readability level of college, high school, middle school, and sixth grade.}
    \label{tab:accuracy_processed}
\end{table}

\paragraph{TIGERScore} We compute TIGERScore \cite{jiang-2024-tigerscore}, which provides explanations in addition to the numeric scores. 
The metric is described by the formula:
\begin{equation}
\{E_1, E_2, \ldots, E_n\} = f(I, x, y')
\end{equation}
where \( f \) is a function that takes the following inputs: \( I \) (instruction), \( x \) (source context), and \( y' \) (system output).
The function \( f \) output a set of structured errors \( \{E_1, E_2, \ldots, E_n\} \). For each error $
E_i = (l_i, a_i, e_i, s_i)$, \( l_i \) denotes the error location, \( a_i \) represents a predefined error aspect, \( e_i \) is a free-text explanation of the error, and \( s_i \) is the score reduction \( \in [-5, -0.5] \) associated with the error.
At the instance level, the overall metric score is the summation of the score reductions for all errors:
$\text{TIGERScore} = \sum_{i=1}^n s_i $.

The native scorer is based on \texttt{Llama-2} \cite{DBLP:journals/corr/abs-2307-09288}. In addition to \texttt{Llama-2}, we send the TIGERScore instructions to the model that performed the task (e.g., \texttt{Mistral-0.2} and \texttt{OpenChat-3.5}), sketching a self-evaluative framework. Through aligning between evaluated and evaluator model, we aim to reduce the negative impacts from hallucination of a single model, i.e., the native \texttt{Llama-2} scorer.
It should nevertheless be noted that this setup may emphasize model biases inherent to the evaluator model \cite{DBLP:conf/nips/PanicksseryBF24}.

\paragraph{BERTScore} As a reference-\underline{based} metric, we parse reference explanations using rule-based methods (App.~\ref{appendix:dataset}) and compute BERTScore \cite{DBLP:conf/iclr/ZhangKWWA20} with end-of-sentence pooling to avoid diluting negations in longer texts.

\paragraph{Human validation} We conduct a human annotation to investigate how human readers view the rationales with distinct readability levels and to validate whether the metric scores could reflect human perception. We choose \textsf{HateXplain} for the setup because it requires little professional knowledge (in comparison to \textsf{HealthFC}) and is performed evenly mediocre across the models, with each of them achieving a similar accuracy score of around 0.5. Using the rationales generated by \texttt{Mistral-0.2} and \texttt{Llama-3} on \textsf{HateXplain}, we sample a split of 200 data points, which consists of 25 random instances per model for each of the four readability levels.


    

We recruit five annotators with computational linguistics and/or machine learning background with at least a Bachelor's degree and have all of them work on the same split. Given the rationales, the annotators are asked to score:
\setlist{nolistsep}
\begin{itemize}[topsep=0pt,leftmargin=*]
    \item \textbf{Readability} (\{30, 50, 70, 90\}): How readable/complex is the generated rationale?
    \item \textbf{Coherence} (4-point Likert scale): To what extent is the rationale logical and reasonable?
    \item \textbf{Informativeness} (4-point Likert): To what extent is the rationale supported by sufficient details?
    \item \textbf{Accuracy} (binary): Does the annotator agree with a prediction after reading the rationale?
\end{itemize}

\section{Results}

We collect predictions and rationales from four models over four datasets (\S \ref{sec:experiments:rationale_generation}). Figure~\ref{fig:output} presents a data instance to exemplify the output of LLM inference as well as each aspect of evaluation. More rationale examples are provided in Appendix~\ref{appendix:sample}.

The four models achieve divergent accuracy scores on the selected tasks (Table~\ref{tab:accuracy_processed}). In most cases, around 5-10\% of instances are unsuccessfully parsed, mostly owing to formatting errors; \texttt{Mistral-0.2} and \texttt{Mixtral-0.1}, however, could hardly follow the instructed output format on particular datasets (\textsf{CAD} and \textsf{HealthFC}), resulting in up to 70\% of instances being removed for these datasets. Since such parsing errors occur only on certain batches, we regard them as special cases similar to those encountered by \citet{tavanaei-2024-solm} and \citet{wu-2024-learning-to-extract} with structured prediction with LLMs. The highest accuracy is reached by \texttt{OpenChat-3.5} for NLI (\textsf{SpanEx}) with a score of 82.1\%. In comparison, multi-class hate speech detection (\textsf{HateXplain}) and medical fact-checking (\textsf{HealthFC}) appear more challenging for all the models, respectively with a peak at 52.0\% (\texttt{OpenChat-3.5}) and 56.4\% (\texttt{Mixtral-0.1}).

Free-text rationales generated under instruction-prompting show a correlative trend in text complexity. Figure~\ref{fig:readability_results} reveals that the requested readability levels introduce notable distinction to text complexity, though the measured output readability may not fully conform with the defined score ranges (Table~\ref{tab:FRE_level}); that is, the distinction is not as significant as the original paradigm. On the other hand, the baseline of \textsf{HealthFC} explanations\footnote{We refer to \textsf{HealthFC} as baseline because the rationales are provided in free-text rather than annotations.} hints a central-leaning tendency for free-text rationales to inherently exhibit medium level readability.

Evaluation with TIGERScore is based on error analyses through score reduction: Each identified error obtains a penalty score (<0), and the entire text is rated the summation of all the reductions. Such design gives 0 to the texts in which no mistake is recognized; in contrast, the more problematic a rationale appears, the lower it scores. In our results (Figure~\ref{fig:tigerscore_results}), we derive non-zero score through further dividing the full-batch score by the amount of non-zero data points, since around half of the rationales are considered fine by the scorer. We also apply the same processing method to self-evaluation with the original model. In most cases, full-batch TIGERScore proportionally decreases along with text complexity, whereas non-zero and self-evaluation do not follow such trend.


In comparison to TIGERScore, BERT similarity provides rather little insight into rationale quality (Appendix~\ref{appendix:results}). Although complex rationales resemble the references more, the correlation between readability and similarity remains weak. Plus, the scores differ more across datasets than across models, making the outcomes less significant.

We conduct a human study (\S \ref{sec:evaluation}) with five annotators, who took around five hours for the 200 samples. 
While calculating agreement, we simplify the results on readability, coherence, and informativeness into two classes owing to the binary nature of 4-point Likert scale; the originally annotated scores are used elsewhere.
We register an agreement of Krippendorff’s \begin{math}\alpha=3.67\%\end{math} and Fleiss’ \begin{math}\kappa=13.92\%\end{math}. Table~\ref{tab:human_annotation} reveals the coherence and informativeness scores. Besides, the human annotators score an accuracy of 23.7\% on recognizing the prompted readability level, while reaching 78.3\% agreement with the model-predicted labels given the rationales.

\begin{figure*}[t]
    \centering
    \resizebox{.95\textwidth}{!}{%
        \includegraphics[width=7.5cm, keepaspectratio]{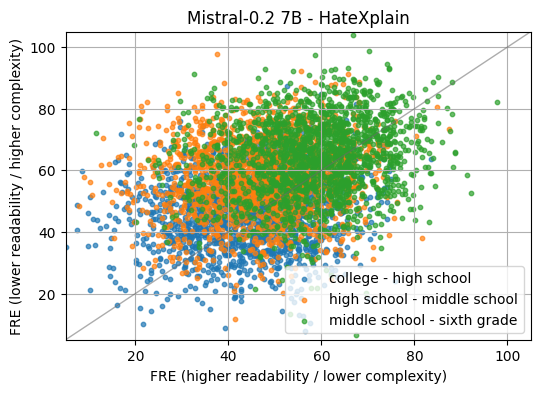}
        \includegraphics[width=7.5cm, keepaspectratio]{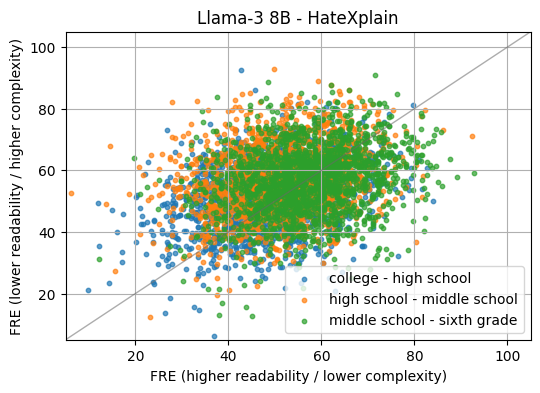}
    }
    \caption{Comparison between FRE scores of two consecutive readability levels. Each dot denotes a data instance, with its more readable rationale positioned on x-axis and less readable on y-axis. The rationales are generated by \texttt{Mistral-0.2} and \texttt{Llama-3} on \textsf{HateXplain}.}
    \label{fig:FRE_HateXplain}
\end{figure*}

\section{Discussions}

Our study aims to respond to three research questions: First, how do LLMs generate different output and free-text rationales under prompted readability level control? Second, how do objective evaluation metrics capture rationale quality of different readability levels? Third, how do human assess the rationales and perceive the NLE outcomes across readability levels?

\subsection{Readability level control under instruction-prompting (RQ1)}

We find free-text rationale generation sensitive to readability level control, whereas the corresponding task predictions remain consistent. This confirms that NLE output is affected by perturbation through instruction prompting.

\begin{table}
    \centering
    \resizebox{\columnwidth}{!}{
    \begin{tabular}{l|cccc|c}
        \multicolumn{6}{c}{\textbf{\textsf{Coherence}}} \\
        \textbf{Readability} & 30 & 50 & 70 & 90 & Avg. \\
        \hline
        \texttt{Mistral-0.2} & 2.84 & 2.98 & \textbf{3.13} & 3.03 & 2.99 \\
        \texttt{Llama-3} & \textbf{3.07} & 3.02 & 2.92 & 2.85 & 2.96 \\
        \hline
        \textbf{Avg.} & 2.96 & 3.00 & \textbf{3.03} & 2.94 & 2.98 \\
        \multicolumn{6}{c}{}
    \end{tabular}}

    \resizebox{\columnwidth}{!}{
    \begin{tabular}{l|cccc|c}
        \multicolumn{6}{c}{\textbf{\textsf{Informativeness}}} \\
        \textbf{Readability} & 30 & 50 & 70 & 90 & Avg. \\
        \hline
        \texttt{Mistral-0.2} & 2.59 & 2.84 & \textbf{3.03} & 2.77 & 2.81 \\
        \texttt{Llama-3} & \textbf{3.02} & 2.93 & 2.86 & 2.86 & 2.92 \\
        \hline
        \textbf{Avg.} & 2.80 & 2.88 & \textbf{2.94} & 2.82 & 2.86 \\
    \end{tabular}}
    \caption{Human-rated scores per model and readability level, with the highest score per model highlighted in bold face. Readability of 30, 50, 70, and 90 respectively refers to the prompted level of college, high school, middle school, and sixth grade.}
    \label{tab:human_annotation}
\end{table}

Without further fine-tuning, the complexity of free-text rationales diverges within a limited range according to readability metrics, showing relative differences rather than precise score mapping. Using \texttt{Mistral-0.2} and \texttt{Llama-3} as examples, Figure~\ref{fig:FRE_HateXplain} plots the distribution of FRE scores between adjacent readability levels. The instances where the model delivers desired readability differentiation fall into the upper-left triangle split by axis \begin{math}y=x\end{math}, while those deviating from the prompted difference appear in the lower-right. The comparison between the two graphs shows that \texttt{Llama-3} aligns the prompted readability level better with generated text complexity, as the distribution area appears more concentrated; meanwhile, \texttt{Mistral-0.2} better differentiates the adjacent readability levels, with more instances falling in the upper-left area.

According to the plots, a considerable amount of rationales nevertheless fail to address the nuances between the prompted levels. This could result from the workflow running through datasets over a given readability level instead of recursively instructing the models to generate consecutive output, i.e., the rationales of different readability levels were generated in several independent sessions. Furthermore, descriptive readability levels do not perfectly match the score ranges shown in Table~\ref{tab:FRE_level}; that is, the two frameworks are only mutually approximate with our experimental setups.

\subsection{Rationale quality presented through metric scores (RQ2)}


We adopt TIGERScore as the main metric for measuring the quality of free-text rationales. On a batch scale, the metric tends to favor rather complex rationales i.e. college or high-school-level. Taking account of the baseline featuring FRE$\approx$50 (Table~\ref{fig:readability_results}), such tendency suggests a slight correspondence between text complexity and explanation quality.

Deriving non-zero scores from full-batch ones, we further find the errors differing in severity at distinct readability levels. After removing error-free instances (where TIGERScore=0), rationales of medium complexity (high school and middle school) can often obtain higher scores. Such divergence implies that less elaborated rationales tend to introduce more mistakes, but they are usually considered minor. In light of both score variations, TIGERScore exhibits characteristics consistent with the central-leaning tendency, i.e., rationales displaying a medium level readability, while potentially echoing the preference for longer texts in LLM-based evaluation \cite{DBLP:journals/corr/abs-2404-04475}.


Full-batch TIGERScore is also found to slightly correlate with task performance (Table~\ref{tab:accuracy_processed}), as better task accuracy usually comes with a higher TIGERScore, though such a tendency doesn't apply across different models. For example, \texttt{Mistral-0.2} achieves better TIGERScore on \textsf{SpanEx} than \texttt{Mixtral-0.1} and \texttt{Llama-3}, whereas both models outperform \texttt{Mistral-0.2} in this task. This could hint at the limitation of the evaluation metric in its nature, as its standard does not unify well across output from different LLMs or tasks.

Other than the reference-free metric, we find BERTScore (Appendix \ref{appendix:results}) differing less significantly, presumably because the meanings of the rationales are mostly preserved across readability levels. Since most reference explanations are parsed under defined rules, such outcome also highlights the gap between rule-based explanations and the actual free-text rationales, signaling linguistic complexity and diversity of explanatory texts.

\subsection{Validation by human annotators (RQ3)}

Our human annotation delivers low agreement scores on the instance level. This results from the designed dimensions aiming for more subjective opinions than a unified standard, capturing human label variation \cite{plank-2022-problem-of-human-label-variation}. Since hate speech fundamentally concerns feelings, agreement scores are typically low. The original labels in \textsf{HateXplain}, for example, reported a Krippendroff’s \begin{math}\alpha=46\%\end{math} \cite{mathew-2021-hatexplain}.

\begin{figure}
    \centering
    \resizebox{.75\columnwidth}{!}{
    \includegraphics{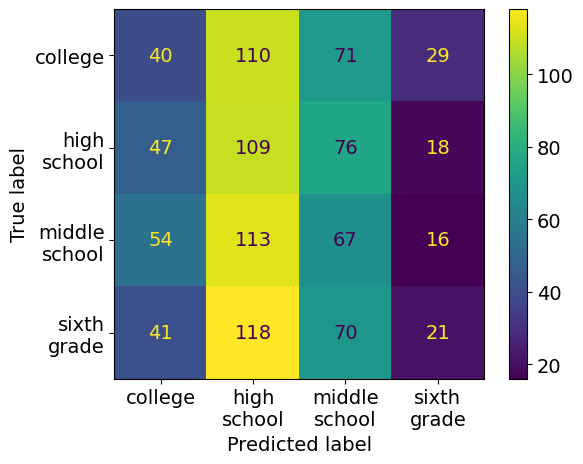}}
    \caption{Human perceived readability level with respect to the prompted ones.}
    \label{fig:human_readability}
\end{figure}

We first discover that human readers do not well perceive the prompted readability levels (Figure~\ref{fig:human_readability}). This corresponds to the misalignment between the prompted levels and the generated rationale complexity. Even so, the rationales receive a generally positive impression (Table~\ref{tab:human_annotation}), with both models scoring significantly above average on a 4-point Likert scale over all the readability levels.

Moreover, the divergence of coherence and informativeness across readability levels (Table~\ref{tab:human_annotation}) shares a similar trend with Figure~\ref{fig:FRE_HateXplain}, with \texttt{Mistral-0.2} having a higher spread than \texttt{Llama-3}, even though the tendency is rarely observed in the other metrics. On one hand, this may imply a gap between metric-captured and human-perceived changes introduced by readability level control; on the other hand, combining these findings, we may also deduce that human readers intrinsically presume free-text rationales to feature a medium level complexity and thereby prefer plain language to unnecessarily complex or over-simplified explanations.

\section{Related Work}

\paragraph{Rationale Evaluation}

Free-text rationale generation was boosted by recent LLMs owing to their capability of explaining their own predictions \cite{DBLP:journals/corr/abs-2401-12874}. Despite lacking a unified paradigm for evaluating rationales, various approaches focused on automatic metrics to minimize human involvement. \begin{math}\nu\end{math}-information \cite{DBLP:conf/emnlp/HewittELM21, DBLP:conf/iclr/XuZSSE20} provided a theoretical basis for metrics such as ReCEval \cite{prasad-2023-receval}, REV \cite{DBLP:conf/acl/ChenB0J0S23}, and RORA \cite{DBLP:conf/acl/JiangLCKDL24}. However, these metrics require training for the scorers to learn new and relevant information with respect to certain tasks.

Alternatively, several studies applied LLMs to perform reference-free evaluation \cite{DBLP:conf/emnlp/LiuIXWXZ23, DBLP:journals/corr/abs-2303-04048}. Similar to TIGERScore \cite{jiang-2024-tigerscore}, InstructScore \cite{DBLP:conf/emnlp/XuWPSFWL23} took advantage of generative models, delivering an reference-free and explainable metric for text generation. However, these approaches could suffer from LLMs’ known problems such as hallucination. As the common methodologies hardly considering both deployment simplicity and assessment accuracy, \citet{DBLP:journals/corr/abs-2401-12874} pointed out the difficulties in designing a paradigm that faithfully reflects the decision-making process of LLMs.

\paragraph{Readability of LLM output}

Rationales generated under readability level control share features similar to those reported by previous studies on NLG-oriented tasks, such as generation of educational texts \cite{DBLP:journals/corr/abs-2406-12787,trott-riviere-2024-gpt-4-readability}, text simplification \cite{DBLP:journals/corr/abs-2409-20246}, and summarization \cite{ribeiro-2023-generating-summaries,wang-demberg-2024-rsa-control}, given that instruction-based methods was proven to alter LLM output in terms of text complexity.
\citet{DBLP:journals/corr/abs-2312-02065} found the readability of LLM output to vary even when controlled through designated prompts. \citet{DBLP:journals/corr/abs-2402-14453} pointed out the limited influence of model parameters on delivering text output of different complexity. While tuning readability remains a significant concern in text simplification and summarization, LLMs were found to tentatively inherit the complexity of input texts and could only rigidly adapt to a broader range of readability \cite{imperial-madabushi-2023-uniform,srikanth-li-2021-elaborative}.

\section{Conclusions}

In this study, we prompted LLMs with distinct readability levels to perturb free-text rationales. We confirmed LLMs' capability of adapting rationales based on instructions, discovering notable shifts in readability with yet a gap between prompted and measured text complexity.
While higher text complexity could sometimes imply better quality, both metric scores and human annotations showed that rationales of approximately high-school complexity were often the most preferred. Moreover, the evaluation outcomes disclosed LLMs' sensitivity to perturbation in rationale generation, potentially supporting a closer connection between NLE and NLG.
Our findings may inspire future works to explore LLMs' explanatory capabilities under perturbation and the application of other NLG-related methodologies to rationale generation.

\section*{Limitations}
\label{sec:limitations}

Owing to time and budget constraints, we are unable to fully explore all the potential variables in the experimental flow, including structuring the prompt, adjusting few-shot training, and instructing different desired output length. Despite the coverage of multiple models and datasets, we only explored the experiments in a single run after trials using web UI. Besides, the occasionally higher ratio of abandoned data instances may induce biases to the demonstrated results; we didn't further probe into the reason for this issue because only particular LLMs have problems on certain datasets, corroborated by concurrent work on structured prediction with LLMs \cite{tavanaei-2024-solm,wu-2024-learning-to-extract}. Lastly, LLM generated text could suffer from hallucination and include false information. Such limitation applies to both rationale generation and LLM-based evaluation.

We were unable to reproduce several NLE-specific metrics. LAS \cite{DBLP:conf/emnlp/HaseZXB20} suffers from outdated library versions, which are no longer available. Although REV \cite{DBLP:conf/acl/ChenB0J0S23} works with the provided toy dataset, we found the implementation fundamentally depending on task-specific data structure, which made it challenging to apply to the datasets we chose. Although we are motivated to conduct perturbation test in an NLG-oriented way, the lack of NLE-specific metrics may limit our insight into the evaluation outcome.

Our human annotators do not share a similar background with the original \textsf{HateXplain} dataset, where the data instances were mostly contributed by North American users. Owing to the different cultural background, biases can be implied and magnified in identifying and interpreting offensive language.

\section*{Ethical Statement}

The datasets of our selection include offensive or hateful contents. Inferring LLM with these materials could result in offensive language usage and even false information involving hateful implications when it comes to hallucination. The human annotators participating in the study were paid at least the minimum wage in conformance with the standards of our host institutions' regions.

\section*{Acknowledgements}
We are indebted to Maximilian Dustin Nasert, Elif Kara, Polina Danilovskaia, and Lin Elias Zander for contributing to the human evaluation.
We thank Leonhard Hennig for his review of our paper draft.
This work has been supported by the German Federal Ministry of Education and Research as part of the project XAINES (01IW20005) and the German Federal Ministry of Research, Technology and Space
as part of the projects VERANDA (16KIS2047) and BIFOLD 24B.

\bibliography{anthology, custom}

\appendix

\section{Data}

\subsection{Task descriptions}
\label{appendix:dataset}

\begin{figure}
   \centering
   \resizebox{.7\columnwidth}{!}{%
        \includegraphics{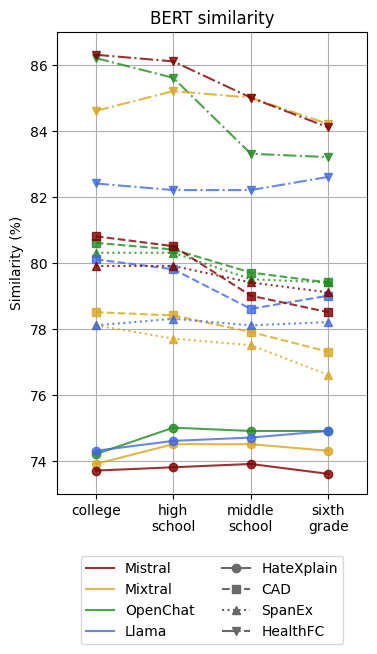}
   }
   \caption{BERTScore similarity between model-generated rationales and reference explanations.}
   \label{fig:similarity_results}
\end{figure}

\begin{table*}
    \centering
    \resizebox{\textwidth}{!}{%
    \begin{tabular}{p{1.75cm}|p{0.75cm}|p{1cm}|p{2.25cm}|p{4.25cm}|p{5cm}}
        \textbf{Dataset} 
            & \textbf{Size}
            & \textbf{\#Test}
            & \textbf{Task}
            & \textbf{Annotations} 
            & \textbf{Sample reference explanation} 
            \\
        \hline
        \multirow{3}{\hsize}{\textsf{HateXplain}} 
            & \multirow{3}{\hsize}{20k}
            & \multirow{3}{\hsize}{1,924}
            & Hate speech classification (multi-class)
            & \multirow{3}{\hsize}{Tokens involving offensive language and their targets} 
            & The text is labeled as hate speech because of expressions against women.
            \\
        \hline
        \multirow{3}{\hsize}{\textsf{CAD}} 
            & \multirow{3}{\hsize}{26k}
            & \multirow{3}{\hsize}{5,307}
            & \multirow{3}{\hsize}{Hate speech detection (binary)}
            & \multirow{3}{\hsize}{Categories of offensive language} 
            & The text is labeled as offensive because the expression involves person directed abuse.      
            \\
        \hline
        \multirow{4}{\hsize}{\textsf{SpanEx}} 
            & \multirow{4}{\hsize}{14k}
            & \multirow{4}{\hsize}{3,865}
            & \multirow{4}{\hsize}{Natural language inference}
            & \multirow{4}{\hsize}{Relevant tokens and their semantic relation} 
            & The relation between hypothesis and premise is contradiction because a girl does not equal to a man.
            \\
        \hline
        \multirow{4}{\hsize}{\textsf{HealthFC}}
            & \multirow{4}{\hsize}{750}
            & \multirow{4}{\hsize}{N/A}
            & \multirow{4}{\hsize}{Fact-checking (multi-class)}
            & \multirow{4}{\hsize}{Excerpts from evidence document that supports or denies the claim (free-text instead of annotations)}
            & There is no scientific evidence that hemolaser treatment has a palliative or curative effect on health problems.
    \end{tabular}
    }
    \caption{Summary of the datasets. Task refers to the adaptation in our experiments instead of the ones proposed by original works. Except for \textsf{HealthFC}, we run the experiments only on test splits.}
    \label{tab:dataset}
\end{table*}

Table~\ref{tab:dataset} summarizes the datasets and the task. Except for \textsf{HealthFC}, every dataset includes explanatory annotations, which are applied to parse reference explanations with rule-based methods. Both aspects are briefly described in Table~\ref{tab:dataset}. The \textsf{HealthFC} dataset excerpts human-written passages as explanations, which are directly adopted as reference rationales in our work.

\begin{figure*}
    \centering
    \resizebox{\textwidth}{!}{\includegraphics{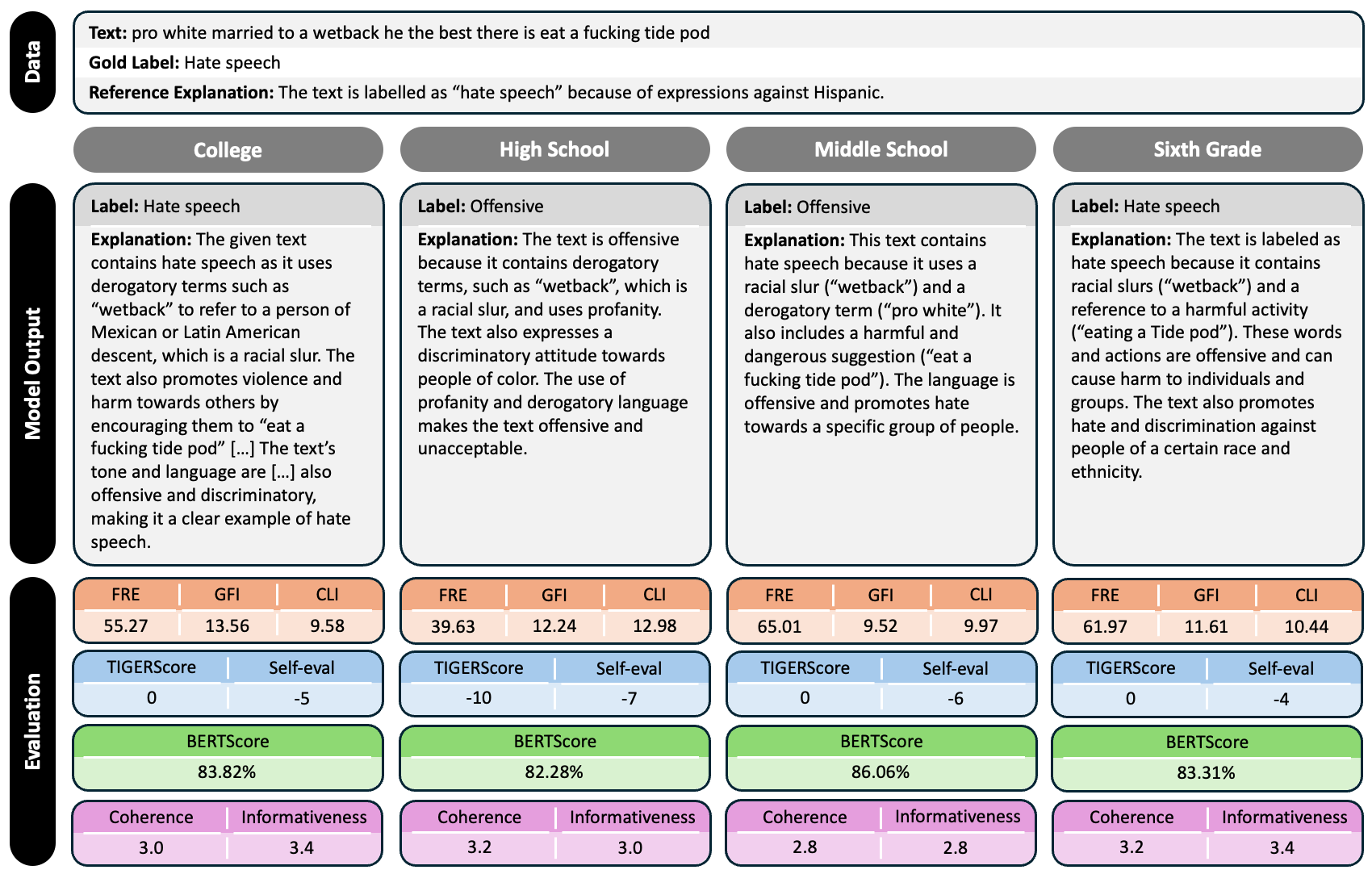}}
    \caption{An example of model predictions and rationales generated by \texttt{Llama-3} on \textsf{HateXplain} along with the evaluation results. Self-eval refers to TIGERScore rated by \texttt{Llama-3}.}
    \label{fig:output_old}
\end{figure*}

\subsection{Sample data instances}
\label{appendix:sample}

Extending Figure~\ref{fig:output}, an additional data point from the \textsf{HateXplain} dataset is provided in Figure~\ref{fig:output_old} to exemplify the scores of human validation.

From Table~\ref{tab:example_hatexplain} to \ref{tab:example_healthfc_neg}, we further provide one data instance for each dataset to exemplify the LLM output under readability level control. Two examples from the \textsf{HealthFC} are given for a more comprehensive comparison between LLM-generated rationales and human-written explanations. In general, although the rationales across readability level tend to appear semantically approximate, they often differ in terms of logical flow and the supporting detail selection, which may imply a strong connection between NLE and NLG, i.e. the generated rationales represent more the learned outcome of LLMs. We also find that the explanations could involve misinterpretation of the context; for example, the high-school-level explanation of \texttt{Mixtral-0.1} on \textsf{HateXplain} (Table~\ref{tab:example_hatexplain}) completely reversed the standpoint of the original text. Furthermore, serious hallucination could occur in the rationale even when the predicted label seems correct. In the high-school-level explanation from \texttt{OpenChat-3.5} on \textsf{CAD} (Table~\ref{tab:example_cad}), ``idiot'' and ``broken in your head'' lead to the \textit{offensive} label, even if these two terms don't really exist in the text; likewise, \texttt{Mistral-0.2} fabricated a digestive condition called ``gossypiasis'' in the sixth-grade-level explanation for \textsf{HealthFC} (Table~\ref{tab:example_healthfc_neg}). Our examples may inspire future works to further investigate perturbed rationale generation.

\section{Metrics for approximating readability}
\label{appendix:metric}

We referred to three metrics to numerically represent text readability. The original formulas of the metrics are listed as below.

Flesch reading ease (FRE) is calculated as follows:

\begin{equation}
FRE = 206.835 - 1.015(w_t/S_t) - 84.6(\sigma_t/w_t)
\end{equation}

\noindent where \begin{math}w_t\end{math} means total words, \begin{math}S_t\end{math} refers to total sentences, and \begin{math}\sigma_t\end{math} represents total syllables.

Gunning fog index (GFI) is based on the formula:

\begin{equation}
GFI = 0.4(w_t/S_t + w_l/S_t)
\end{equation}

\noindent where \begin{math}w_t\end{math} represents total words, and \begin{math}S_t\end{math} means total sentences. \begin{math}w_l\end{math} is the amount of long words that consists of more than seven alphabets.

The formula of Coleman-Liau index (CLI) goes as follows:

\begin{equation}
CLI = 0.0588\bar{L} - 0.296\bar{S} - 15.8
\end{equation}

\noindent where \begin{math}\bar{L}\end{math} describes the average number of letters every 100 words, and \begin{math}\bar{S}\end{math} represents the average amount of sentences every 100 words.

\section{Raw evaluation data of model predictions and rationales}

\label{appendix:results}

The appended tables include the raw data presented in the paper as processed results or graphs. Table~\ref{tab:accuracy_raw} denotes task accuracy scores without removing unsuccessfully parsed data instances; that is, in contrast to Table~\ref{tab:accuracy_processed}, instances with empty prediction are considered incorrect here.

Table~\ref{tab:FRE}, \ref{tab:GFI}, and \ref{tab:CLI} respectively include the three readability scores over each batch, which are visualised in Figure~\ref{fig:tigerscore_results}. Table~\ref{tab:TIGERScore} provides the detailed numbers shown in Figure~\ref{fig:tigerscore_results}. Figure~\ref{fig:similarity_results} visualizes the similarity scores, with the exact numbers described in Table~\ref{tab:BERT}. The figure shows that the scores show rather little variation, with only minor differences in similarity scores within the same task. On one hand, such outcome implies that meanings of the rationales are mostly preserved across readability levels; on the other hand, this may reflect the constraints of both BERT measuring similarity, given that cosine similarity tends to range between 0.6 and 0.9, and parsing reference explanations out of fixed rules, which fundamentally limits the lexical complexity of the standard being used.

In every table, readability of 30, 50, 70, and 90 respectively refers to the prompted readability level of college, high school, middle school, and sixth grade.


\begin{table}
    \centering
    \renewcommand*{\arraystretch}{}
    \resizebox{.9\columnwidth}{!}{%
    \begin{tabular}{ll|cccc}
        & \textbf{Readability} & 30 & 50 & 70 & 90 \\
        \hline
        \parbox[t]{2mm}{\multirow{4}{*}{\rotatebox[origin=c]{90}{\small{\textbf{\textsf{HateXplain}}}}}} & \texttt{Mistral-0.2} & 48.1 & 48.2 & 51.5 & 50.9 \\
        & \texttt{Mixtral-0.1} & 41.7 & 42.5 & 42.1 & 42.7 \\
        & \texttt{OpenChat-3.5} & 50.2 & 50.3 & \textbf{52.0} & 49.5 \\
        & \texttt{Llama-3} & 50.2 & 50.8* & 50.0 & 49.5 \\
        \hline
        \parbox[t]{2mm}{\multirow{4}{*}{\rotatebox[origin=c]{90}{\small{\textbf{\textsf{CAD}}}}}} & \texttt{Mistral-0.2} & \textbf{81.3}* & 81.1 & 78.7 & 76.6 \\
        & \texttt{Mixtral-0.1} & 60.8* & 59.6 & 59.2 & 57.9 \\
        & \texttt{OpenChat-3.5} & 74.4 & 75.4 & 74.6 & 74.6 \\
        & \texttt{Llama-3} & 48.1 & 46.2 & 44.7 & 43.5 \\
        \hline
        \parbox[t]{2mm}{\multirow{4}{*}{\rotatebox[origin=c]{90}{\small{\textbf{\textsf{SpanEx}}}}}} & \texttt{Mistral-0.2} & 33.9 & 34.6 & 35.8 & 36.1 \\
        & \texttt{Mixtral-0.1} & 53.1 & 50.1 & 50.5 & 53.2 \\
        & \texttt{OpenChat-3.5} & 81.8 & \textbf{82.1}* & 81.4 & 82.0 \\
        & \texttt{Llama-3} & 40.0 & 38.0 & 36.8 & 36.8 \\
        \hline
        \parbox[t]{2mm}{\multirow{4}{*}{\rotatebox[origin=c]{90}{\small{\textbf{\textsf{HealthFC}}}}}} & \texttt{Mistral-0.2} & \textbf{50.4} & 49.3 & 50.4 & 47.8 \\
        & \texttt{Mixtral-0.1} & 46.8 & 48.0 & 46.9 & 49.0 \\
        & \texttt{OpenChat-3.5} & 48.9 & 49.7 & 49.7 & 49.5 \\
        & \texttt{Llama-3} & 26.9 & 29.2 & 28.2 & 25.7 \\
    \end{tabular}
    }
    \caption{Raw task accuracy scores (\%), in which unsuccessfully parsed model output were considered incorrect. The best score(s) achieved by a model are starred, and best accuracy per task are highlighted in bold face.}
    \label{tab:accuracy_raw}
\end{table}


\begin{table}
    \centering
    \renewcommand*{\arraystretch}{}
    \resizebox{.9\columnwidth}{!}{%
    \begin{tabular}{ll|cccc}
        & \textbf{Readability} & 30 & 50 & 70 & 90 \\
        \hline
        \parbox[t]{2mm}{\multirow{4}{*}{\rotatebox[origin=c]{90}{\small{\textbf{\textsf{HateXplain}}}}}} & \texttt{Mistral-0.2} & 48.1 & 50.9 & 56.6 & 62.1 \\
        & \texttt{Mixtral-0.1} & 44.8 & 47.2 & 58.0 & 64.0 \\
        & \texttt{OpenChat-3.5} & 50.7 & 54.9 & 62.0 & 64.1 \\
        & \texttt{Llama-3} & 49.1 & 51.5 & 57.0 & 56.8 \\
        \hline
        \parbox[t]{2mm}{\multirow{4}{*}{\rotatebox[origin=c]{90}{\small{\textbf{\textsf{CAD}}}}}} & \texttt{Mistral-0.2} & 45.8 & 47.8 & 56.5 & 59.9 \\
        & \texttt{Mixtral-0.1} & 48.0 & 49.9 & 55.5 & 59.0 \\
        & \texttt{OpenChat-3.5} & 53.3 & 56.1 & 61.6 & 63.1 \\
        & \texttt{Llama-3} & 47.1 & 50.0 & 55.5 & 54.6 \\
        \hline
        \parbox[t]{2mm}{\multirow{4}{*}{\rotatebox[origin=c]{90}{\small{\textbf{\textsf{SpanEx}}}}}} & \texttt{Mistral-0.2} & 52.0 & 54.4 & 60.0 & 62.1 \\
        & \texttt{Mixtral-0.1} & 59.5 & 61.4 & 66.9 & 71.8 \\
        & \texttt{OpenChat-3.5} & 61.3 & 66.8 & 73.3 & 73.8 \\
        & \texttt{Llama-3} & 51.1 & 55.0 & 59.7 & 62.0 \\
        \hline
        \parbox[t]{2mm}{\multirow{4}{*}{\rotatebox[origin=c]{90}{\small{\textbf{\textsf{HealthFC}}}}}} & \texttt{Mistral-0.2} & 44.2 & 44.2 & 47.5 & 48.8 \\
        & \texttt{Mixtral-0.1} & 41.3 & 44.0 & 51.7 & 56.2 \\
        & \texttt{OpenChat-3.5} & 43.8 & 51.1 & 62.8 & 63.8 \\
        & \texttt{Llama-3} & 41.2 & 44.2 & 47.5 & 48.8 \\
    \end{tabular}
    }
    \caption{FRE scores of model-generated rationales.}
    \label{tab:FRE}
\end{table}


\begin{table}
    \centering
    \renewcommand*{\arraystretch}{}
    \resizebox{.9\columnwidth}{!}{%
    \begin{tabular}{ll|cccc}
        & \textbf{Readability} & 30 & 50 & 70 & 90 \\
        \hline
        \parbox[t]{2mm}{\multirow{4}{*}{\rotatebox[origin=c]{90}{\small{\textbf{\textsf{HateXplain}}}}}} & \texttt{Mistral-0.2} & 14.2 & 13.6 & 12.2 & 11.2 \\
        & \texttt{Mixtral-0.1} & 15.1 & 14.5 & 12.0 & 10.7 \\
        & \texttt{OpenChat-3.5} & 13.6 & 12.8 & 11.4 & 10.9 \\
        & \texttt{Llama-3} & 13.9 & 13.4 & 12.3 & 12.3 \\
        \hline
        \parbox[t]{2mm}{\multirow{4}{*}{\rotatebox[origin=c]{90}{\small{\textbf{\textsf{CAD}}}}}} & \texttt{Mistral-0.2} & 14.8 & 14.3 & 12.2 & 11.5 \\
        & \texttt{Mixtral-0.1} & 14.1 & 13.6 & 12.4 & 11.7 \\
        & \texttt{OpenChat-3.5} & 12.9 & 12.3 & 11.2 & 10.9 \\
        & \texttt{Llama-3} & 14.1 & 13.3 & 12.1 & 12.3 \\
        \hline
        \parbox[t]{2mm}{\multirow{4}{*}{\rotatebox[origin=c]{90}{\small{\textbf{\textsf{SpanEx}}}}}} & \texttt{Mistral-0.2} & 12.7 & 12.1 & 11.1 & 10.8 \\
        & \texttt{Mixtral-0.1} & 11.8 & 11.6 & 10.3 & 9.5 \\
        & \texttt{OpenChat-3.5} & 10.7 & 9.9 & 9.0 & 8.9 \\
        & \texttt{Llama-3} & 13.2 & 12.3 & 11.2 & 10.8 \\
        \hline
        \parbox[t]{2mm}{\multirow{4}{*}{\rotatebox[origin=c]{90}{\small{\textbf{\textsf{HealthFC}}}}}} & \texttt{Mistral-0.2} & 15.1 & 14.2 & 13.4 & 13.2 \\
        & \texttt{Mixtral-0.1} & 14.3 & 14.0 & 12.5 & 11.7 \\
        & \texttt{OpenChat-3.5} & 13.6 & 12.3 & 10.5 & 10.1 \\
        & \texttt{Llama-3} & 15.1 & 14.2 & 13.4 & 13.2 \\
    \end{tabular}
    }
    \caption{GFI scores of model-generated rationales.}
    \label{tab:GFI}
\end{table}


\begin{table}
    \centering
    \renewcommand*{\arraystretch}{}
    \resizebox{.9\columnwidth}{!}{%
    \begin{tabular}{ll|cccc}
        & \textbf{Readability} & 30 & 50 & 70 & 90 \\
        \hline
        \parbox[t]{2mm}{\multirow{4}{*}{\rotatebox[origin=c]{90}{\small{\textbf{\textsf{HateXplain}}}}}} & \texttt{Mistral-0.2} & 12.2 & 11.7 & 10.8 & 9.8 \\
        & \texttt{Mixtral-0.1} & 12.7 & 12.4 & 10.7 & 9.7 \\
        & \texttt{OpenChat-3.5} & 11.8 & 11.2 & 10.0 & 9.5 \\
        & \texttt{Llama-3} & 12.0 & 11.5 & 10.7 & 10.7 \\
        \hline
        \parbox[t]{2mm}{\multirow{4}{*}{\rotatebox[origin=c]{90}{\small{\textbf{\textsf{CAD}}}}}} & \texttt{Mistral-0.2} & 12.5 & 12.2 & 11.0 & 10.5 \\
        & \texttt{Mixtral-0.1} & 12.1 & 11.8 & 11.0 & 10.4 \\
        & \texttt{OpenChat-3.5} & 11.0 & 10.6 & 9.7 & 9.4 \\
        & \texttt{Llama-3} & 12.2 & 11.9 & 11.0 & 11.1 \\
        \hline
        \parbox[t]{2mm}{\multirow{4}{*}{\rotatebox[origin=c]{90}{\small{\textbf{\textsf{SpanEx}}}}}} & \texttt{Mistral-0.2} & 11.6 & 11.2 & 10.2 & 9.8 \\
        & \texttt{Mixtral-0.1} & 10.5 & 10.1 & 9.2 & 8.1 \\
        & \texttt{OpenChat-3.5} & 11.0 & 9.8 & 8.1 & 8.1 \\
        & \texttt{Llama-3} & 11.9 & 11.5 & 10.7 & 10.4 \\
        \hline
        \parbox[t]{2mm}{\multirow{4}{*}{\rotatebox[origin=c]{90}{\small{\textbf{\textsf{HealthFC}}}}}} & \texttt{Mistral-0.2} & 13.8 & 13.2 & 12.8 & 12.1 \\
        & \texttt{Mixtral-0.1} & 14.2 & 13.9 & 12.6 & 11.8 \\
        & \texttt{OpenChat-3.5} & 14.0 & 12.7 & 10.5 & 10.4 \\
        & \texttt{Llama-3} & 13.8 & 13.2 & 12.8 & 12.6 \\
    \end{tabular}
    }
    \caption{CLI scores of model-generated rationales.}
    \label{tab:CLI}
\end{table}

\begin{table}
    \centering
    \small
    \resizebox{.9\columnwidth}{!}{
    \begin{tabular}{l|cccc}
        \multicolumn{5}{c}{\textbf{\textsf{HateXplain}}} \\
        \textbf{Readability} & 30 & 50 & 70 & 90 \\
        \hline
         & \textbf{-3.15} & -3.25 & -3.73 & -3.93 \\
        \texttt{Mistral-0.2} & 648 & 679 & 784 & \underline{822} \\
         & -9.10 & -8.99 & -8.90* & -8.99 \\
        \hline
         & \textbf{-3.44} & -3.68 & -3.82 & -4.48 \\
        \texttt{Mixtral-0.1} & 750 & 747 & 782 & \underline{882} \\
         & -7.95* & -8.30 & -8.34 & -8.73 \\
        \hline
         & \textbf{-3.62} & -3.88 & -4.24 & -4.31 \\
        \texttt{OpenChat-3.5} & 860 & 966 & \underline{1,067} & 1,044 \\
         & -7.85 & -7.53 & -7.47* & -7.77 \\
        \hline
         & \textbf{-3.41} & -3.74 & -3.90 & -4.03 \\
        \texttt{Llama-3} & 701 & 737 & \underline{808} & 782 \\
         & -9.27 & -9.62 & -9.16* & -9.73 \\
        \multicolumn{5}{c}{}
    \end{tabular}}

    \resizebox{.9\columnwidth}{!}{
    \begin{tabular}{l|cccc}
        \multicolumn{5}{c}{\textbf{\textsf{CAD}}} \\
        \textbf{Readability} & 30 & 50 & 70 & 90 \\
        \hline
         & \textbf{-1.79} & -1.91 & -2.53 & -2.71 \\
        \texttt{Mistral-0.2} & 1,135 & 1,216 & 1,688 & \underline{1,768} \\
         & -8.14 & -8.15 & -7.74* & -7.87 \\
        \hline
         & \textbf{-2.27} & -2.30 & -2.77 & -3.21 \\
        \texttt{Mixtral-0.1} & 1,471 & 1,477 & 1,786 & \underline{1,989} \\
         & -7.57* & -7.59 & -7.63 & 7.97 \\
        \hline
         & -2.30 & \textbf{-2.29} & -2.57 & -2.86 \\
        \texttt{OpenChat-3.5} & 1,427 & 1,468 & 1,652 & \underline{1,769} \\
         & -8.23 & -7.98 & -7.90* & -8.30 \\
        \hline
         & \textbf{-3.04} & -3.58 & -4.17 & -4.52 \\
        \texttt{Llama-3} & 1,399 & 1,557 & 1,747 & \underline{1,774} \\
         & -9.16* & -9.59 & -9.77 & -10.59 \\
        \multicolumn{5}{c}{}
    \end{tabular}}

    \resizebox{.9\columnwidth}{!}{
    \begin{tabular}{l|cccc}
        \multicolumn{5}{c}{\textbf{\textsf{SpanEx}}} \\
        \textbf{Readability} & 30 & 50 & 70 & 90 \\
        \hline
         & \textbf{-2.76} & -2.88 & -3.31 & -3.52 \\
        \texttt{Mistral-0.2} & 1,193 & 1,235 & 1,472 & \underline{1,479} \\
         & -8.64 & -8.75 & -8.51* & -8.90 \\
        \hline
         & -3.29 & \textbf{-3.28} & -3.82 & -4.42 \\
        \texttt{Mixtral-0.1} & 1,552 & 1,578 & 1,820 & \underline{1,994} \\
         & -7.43 & -7.18* & -7.41 & -7.83 \\
        \hline
         & \textbf{-1.85} & -2.18 & -2.95 & -3.18 \\
        \texttt{OpenChat-3.5} & 916 & 991 & 1,299 & \underline{1,322} \\
         & -7.45* & -7.98 & -8.30 & -8.88 \\
        \hline
         & \textbf{-3.86} & -4.48 & -5.25 & -5.41 \\
        \texttt{Llama-3} & 1,500 & 1,714 & 1,914 & \underline{1,926} \\
         & -9.25 & -9.19* & -9.31 & -9.71 \\
         \multicolumn{5}{c}{}
    \end{tabular}}

    \resizebox{.9\columnwidth}{!}{
    \begin{tabular}{l|cccc}
        \multicolumn{5}{c}{\textbf{\textsf{HealthFC}}} \\
        \textbf{Readability} & 30 & 50 & 70 & 90 \\
        \hline
         & -1.20 & \textbf{-0.94} & -1.07 & -1.11 \\
        \texttt{Mistral-0.2} & 169 & 165 & 158 & \underline{179} \\
         & -5.09 & -4.02* & -4.83 & -4.49 \\
        \hline
         & -1.96 & \textbf{-1.72} & -2.01 & -2.16 \\
        \texttt{Mixtral-0.1} & 246 & 236 & 238 & \underline{256} \\
         & -5.11 & -4.67* & -5.42 & -5.53 \\
        \hline
         & \textbf{-3.15} & -3.28 & -3.80 & -4.10 \\
        \texttt{OpenChat-3.5} & 380 & 362 & 397 & \underline{411} \\
         & -5.86* & -6.34 & -6.73 & -7.10 \\
        \hline
         & -6.49 & \textbf{-6.39} & -6.77 & -6.99 \\
        \texttt{Llama-3} & \underline{513} & 484 & 497 & 496 \\
         & -9.08* & -9.32 & -9.55 & -9.73 \\
    \end{tabular}}

    \caption{TIGERScore of the model-generated rationales. For each model, the first score is full-batch TIGERScore, which averages among all instances. The second number denotes the number of non-zero instances, and the third row shows non-zero TIGERScore, where instances scoring 0 were removed. Bold font highlights the best full-batch scores. The highest amount of non-zero instances are underlines. And the best non-zero scores are starred.}
    \label{tab:TIGERScore}
\end{table}

\section{Human annotation guidelines}
\label{appendix:guideline}

Table~\ref{tab:guideline} presents the annotation guidelines, which describe the four aspects that were to be annotated. We assigned separate Google spreadsheets to the recruited annotators as individual workspace. In the worksheet, 20 annotated instances were provided as further examples along with a brief description of the workflow.

\begin{table}
    \centering
    \resizebox{\columnwidth}{!}{
    \begin{tabular}{l|cccc}
        \multicolumn{5}{c}{\textbf{\textsf{HateXplain}}} \\
        \textbf{Readability} & 30 & 50 & 70 & 90 \\
        \hline
        \texttt{Mistral-0.2} & 73.7 & 73.8 & 73.9* & 73.6 \\
        \texttt{Mixtral-0.1} & 73.9 & 74.5* & 74.5* & 74.3 \\
        \texttt{OpenChat-3.5} & 74.2 & \textbf{75.0}* & 74.9 & 74.9 \\
        \texttt{Llama-3} & 74.3 & 74.6 & 74.7 & 74.9* \\
        \multicolumn{5}{c}{}
    \end{tabular}}

    \resizebox{\columnwidth}{!}{
    \begin{tabular}{l|cccc}
        \multicolumn{5}{c}{\textbf{\textsf{CAD}}} \\
        \textbf{Readability} & 30 & 50 & 70 & 90 \\
        \hline
        \texttt{Mistral-0.2} & \textbf{80.8}* & 80.5 & 79.0 & 78.5 \\
        \texttt{Mixtral-0.1} & 78.5* & 78.4 & 77.9 & 77.3 \\
        \texttt{OpenChat-3.5} & 80.6* & 80.4 & 79.7 & 79.4 \\
        \texttt{Llama-3} & 80.1* & 79.8 & 78.6 & 79.0 \\
        \multicolumn{5}{c}{}
    \end{tabular}}

    \resizebox{\columnwidth}{!}{
    \begin{tabular}{l|cccc}
        \multicolumn{5}{c}{\textbf{\textsf{SpanEx}}} \\
        \textbf{Readability} & 30 & 50 & 70 & 90 \\
        \hline
        \texttt{Mistral-0.2} & 79.9* & 79.9* & 79.4 & 79.1 \\
        \texttt{Mixtral-0.1} & 78.1* & 77.7 & 77.5 & 76.6 \\
        \texttt{OpenChat-3.5} & \textbf{80.3}* & \textbf{80.3}* & 79.5 & 79.4 \\
        \texttt{Llama-3} & 78.1 & 78.3* & 78.1 & 78.2 \\
        \multicolumn{5}{c}{}
    \end{tabular}}

    \resizebox{\columnwidth}{!}{
    \begin{tabular}{l|cccc}
        \multicolumn{5}{c}{\textbf{\textsf{HealthFC}}} \\
        \textbf{Readability} & 30 & 50 & 70 & 90 \\
        \hline
        \texttt{Mistral-0.2} & 86.3* & 86.1 & 85.0 & 84.1 \\
        \texttt{Mixtral-0.1} & 84.6 & 85.2* & 85.0 & 84.2 \\
        \texttt{OpenChat-3.5} & \textbf{86.2}* & 85.6 & 83.3 & 83.2 \\
        \texttt{Llama-3} & 82.4 & 82.2 & 82.3 & 82.6*
    \end{tabular}}
    
    \caption{BERT similarity scores between rationale and reference explanation (\%). For each task, star sign marks out the best score(s) achieved by each model, and bold font highlights the task-specific highest score.}
    \label{tab:BERT}
\end{table}

\begin{landscape}
    \begin{table}[]
    \small
        \centering
        \begin{tabularx}{1.6\textwidth}{>{\hsize=.08\hsize\linewidth=\hsize}X|>{\hsize=.25\hsize\linewidth=\hsize}X|>{\hsize=.25\hsize\linewidth=\hsize}X|>    {\hsize=.25\hsize\linewidth=\hsize}X|>{\hsize=.25\hsize\linewidth=\hsize}X}
            \textbf{Model} & \texttt{Mistral-0.2} & \texttt{Mixtral-0.1} & \texttt{OpenChat-3.5} & \texttt{Llama-3} \\
            \hline
            \textbf{Text} & \multicolumn{4}{X}{<user> ccp china z i is happy to kill all freedom fighter please pass hk human rights democracy act and free hong kong standwith hk} \\
            \hline
            \textbf{Gold label} & \multicolumn{4}{X}{normal} \\
            \hline
            \textbf{Reference explanation} & \multicolumn{4}{X}{The text is labelled as ``normal.''} \\
            \hline
            \textbf{Model-generated answer and rationale}
            & \textbf{College:} \textit{Normal.} The text does not contain any explicit hate speech or offensive language against any specific target group. It is simply an expression of support for human rights and democracy in Hong Kong
            \newline \textbf{High School:} \textit{Normal.} This text is expressing support for a political cause, the democratization of Hong Kong, and advocacy for a US law. The user is not making negative or hateful comments, so the label is normal. The user's intention is to raise awareness and encourage action for a human rightsissue.
            \newline \textbf{Middle School:} \textit{Normal.} The user is expressing their opinion in support of human rights and democracy for Hong Kong, using terms like ``freedom fighters'' and ``human rights'' that are common when talking about politics and social issues. This is considered a normal expression of opinion and does not target any specific group with hate speech.
            \newline \textbf{Sixth Grade:} \textit{Normal.} This text is not offensive or hateful. It's a user expressing their support for human rights and democracy in Hong Kong. They are asking for the passing of the ``Hong Kong Human Rights and Democracy Act'' to help those fighting for freedom in China.
            & \textbf{College:} \textit{Normal.}
            The text is presenting an opinion about China, without offending any group of people or inciting hate speech. The user's stance on the issue is clear, but it doesn't break any of the community guidelines.
            \newline \textbf{High School:} \textit{Offensive.} The user supports the CCP in China, which is a political party that has been responsible for human rights violations. The user also seems to show no concern for innocent civilians who may be caught in crossfire between the CCP and ``freedom fighters.'' These actions or perspectives can be considered offensive because they are harmful towards innocent individuals and communities.
            \newline \textbf{Middle School:} \textit{Normal.} The user expresses support for Hong Kong's human rights, democracy, and freedom. The user does not offend or hate any target group, so the label is ``normal.''
            \newline \textbf{Sixth Grade:} \textit{Normal.} The user is expressing a political opinion and showing support for a cause. They are not insulting or attacking any group of people. They are simply stating their political beliefs and hoping for a change in the politics of China.
            & \textbf{College:} \textbf{College:} \textit{Offensive.} The text is offensive because it expresses a strong negative sentiment towards a political group (CCP China) and uses the phrase ``happy to kill all freedom fighters'' which is a violent and aggressive statement. This type of language may incite negative feelings and further divide the community.
            \newline \textbf{High School:} \textit{Offensive.} The text is offensive because it uses strong language to express a political opinion and it could be hurtful to some people. The text also uses the term ``ccp china z i'' which could be seen as a derogatory term.
            \newline \textbf{Middle School:} \textit{Hate speech.} The text is labeled as hate speech because it expresses strong negative emotions towards a specific group (CCP China) and seems to encourage violence against them. This is not appropriate and may cause harm to the targeted group.
            \newline \textbf{Sixth Grade:} \textit{Hate speech.} The text is hate speech because it targets a specific group of people, the Chinese Communist Party, and expresses a desire for them to be killed. This is not a normal or offensive statement, as it directly incites violence against a group of people.
            & \textbf{College:} \textbf{College:} \textit{Offensive.} The text is labeled as offensive because it promotes violence and harm towards a group of people (freedom fighters) based on their political beliefs. The text also uses aggressive language, such as ``kill all,'' which can be seen as threatening and harmful. Additionally, the text makes a generalization about a group of people (Chinese) based on their national identity, which can be seen as discriminatory.
            \newline \textbf{High School:} \textit{Hate speech.} This text is classified as hate speech because it openly supports the killing of freedom fighters [...]. It also uses a derogatory term (``kill all'') which is a clear indicator of hate speech. The text is not only offensive but also promotes harm and violence against a group of people, which is a characteristic of hate speech.
            \newline \textbf{Middle School:} \textit{Offensive.} This text is offensive because it contains hate speech towards a specific group, CCP China. The user is calling for the killing of ``freedom fighters,'' which could be interpreted as a call to violence against a particular group of people. This language is harmful and unacceptable.
            \newline \textbf{Sixth Grade:} \textit{Offensive.} This text is offensive because it uses a derogatory term ``freedom fighter'' to describe people who are fighting for their rights and freedoms. The text also calls for violence against the people of Hong Kong, which is a violation of human rights. The language used is aggressive and threatening, and it is likely to cause harm to individuals and communities.
        \end{tabularx}
        \caption{An example data instance from the \textsf{HateXplain} dataset. Owing to the limited space, some longer rationales are partially omitted and indicated with [...].}
        \label{tab:example_hatexplain}
    \end{table}
\end{landscape}

\begin{landscape}
    \begin{table}[]
    \small
        \centering
        \begin{tabularx}{1.6\textwidth}{>{\hsize=.08\hsize\linewidth=\hsize}X|>{\hsize=.25\hsize\linewidth=\hsize}X|>{\hsize=.25\hsize\linewidth=\hsize}X|>    {\hsize=.25\hsize\linewidth=\hsize}X|>{\hsize=.25\hsize\linewidth=\hsize}X}
            \textbf{Model} & \texttt{Mistral-0.2} & \texttt{Mixtral-0.1} & \texttt{OpenChat-3.5} & \texttt{Llama-3} \\
            \hline
            \textbf{Text} & \multicolumn{4}{X}{You don't even know what real communism is you nazi} \\
            \hline
            \textbf{Gold label} & \multicolumn{4}{X}{offensive} \\
            \hline
            \textbf{Reference explanation} & \multicolumn{4}{X}{The text is labelled as offensive because the expression involves person directed abuse.} \\
            \hline
            \textbf{Model-generated answer and rationale}
            & \textbf{College:} \textit{Offensive.} This statement is offensive because it uses a derogatory term, ``you Nazi,'' to attack or belittle someone without providing any reasons or evidence to support the claim. The term ``Nazi'' is often used to stigmatize people with extreme political beliefs, but using it randomly and without context in a personal attack is not productive or respectful.
            \newline \textbf{High School:} \textit{Offensive.} The text uses the word ``nazi'' as an insult towards someone, and it's not related to the topic of the text. This can be considered offensive as it's an attempt to belittle or humiliate someone based on their beliefs or background, which is not respectful or acceptable in a civil discourse.
            \newline \textbf{Middle School:} \textit{Offensive.} This label is for texts that are hurtful or mean towards a group of people. In this case, the person is using a derogatory term ``nazi'' towards people who don't agree with their view on communism. It is important to remember that using hurtful words can be harmful and disrespectful to others.
            \newline \textbf{Sixth Grade:} \textit{Offensive.} This text is offensive because it calls someone a ``nazi'' without any reason, which is a hurtful word. Using words like that to insult someone is not kind or respectful, and it goes against our rule of treating everyone with kindness and respect, no matter what they believe or say.
            & \textbf{College:} \textit{Offensive.} The text is offensive because it labels someone as a ``nazi.'' The term ``nazi'' is often times used as an insult to diminish one's point of view and discredit their opinions. This type of language aims to intimidate or provoke, causing an unsafe environment for open communication.
            \newline \textbf{High School:} \textit{Offensive.} This text is offensive because it uses an offensive term ``nazi'' to attack someone, suggesting they are ignorant or stupid (as shown in the phrase ``You don't even know what real communism is''). Using an offensive term to insult someone is not respectful and therefore it count as offensive.
            \newline \textbf{Middle School:} \textit{Offensive.} The text uses the term ``nazi'' as an insult, which is a derogatory term for people who hold views or characteristics similar to Nazis. In this case, the user is likely not referring to actual Nazis, but is using the term as an insult to criticize someone's views on communism. Using a term as an insult shows a lack of respect and is therefore considered offensive.
            \newline \textbf{Sixth Grade:} \textit{Offensive.} The sentence uses an offensive term called ``nazi'' to insult someone who does not understand what real communism is. Comparing someone to a ``nazi'' is a form of name-calling and can be hurtful and offensive. Just like in the second example, using offensive language is not respectful or appropriate, and it violates the community guidelines.
            & \textbf{College:} \textit{Offensive.} This text is offensive because it uses derogatory language to refer to someone's political beliefs. The use of ``nazi'' as an insult is not suitable for a respectful conversation.
            \newline \textbf{High School:} \textit{Offensive.} The text uses derogatory terms like ``idiot'' and ``broken in your head'' to belittle and insult someone's intelligence. This kind of language can be hurtful and offensive to others.
            \newline \textbf{Middle School:} \textit{Offensive.} This text uses strong language and may be hurtful to some people. It is not suitable for a friendly conversation. It is important to be kind and respectful to others.
            \newline \textbf{Sixth Grade:} \textit{Offensive.} The given text uses strong language and might hurt someone's feelings. It's not nice to call people names or make fun of their ideas.
            & \textbf{College:} \textit{Offensive.} The text is offensive because it uses a derogatory term and makes an inflammatory comparison between communism and Nazism, which is a sensitive and divisive topic. The use of the word ``nazi'' as an insult is a form of hate speech and is likely to offend people who identify as Jewish or have been affected by the Holocaust. The text also lacks any constructive or respectful language, and instead resorts to name-calling and personal attacks.
            \newline \textbf{High School:} \textit{Offensive.} The text contains a derogatory term (``nazi'') and insults (``You don't even know what real communism is'') which can be hurtful and offensive to some people.
            \newline \textbf{Middle School:} \textit{Offensive.} The text uses profanity and a derogatory term (``nazi'') which is offensive and may be hurtful to people who identify with the targeted group. The use of the term ``nazi'' is also a stereotype and can be seen as a form of hate speech. The text also contains a threat or insult, making it even more offensive.
            \newline \textbf{Sixth Grade:} \textit{Offensive.} This text is offensive because it uses a racial slur ``nazi'' to describe communism, which is a derogatory and hurtful term. It's also an example of hate speech, which can cause harm to people who identify with the term. Using this kind of language is not acceptable and can create a hostile environment.
        \end{tabularx}
        \caption{An example data instance from the \textsf{CAD} dataset.}
        \label{tab:example_cad}
    \end{table}
\end{landscape}

\begin{landscape}
    \begin{table}[]
    \small
        \centering
        \begin{tabularx}{1.6\textwidth}{>{\hsize=.08\hsize\linewidth=\hsize}X|>{\hsize=.25\hsize\linewidth=\hsize}X|>{\hsize=.25\hsize\linewidth=\hsize}X|>    {\hsize=.25\hsize\linewidth=\hsize}X|>{\hsize=.25\hsize\linewidth=\hsize}X}
            \textbf{Model} & \texttt{Mistral-0.2} & \texttt{Mixtral-0.1} & \texttt{OpenChat-3.5} & \texttt{Llama-3} \\
            \hline
            \textbf{Premise} & \multicolumn{4}{X}{People jump over a mountain crevasse on a rope.} \\
            \textbf{Hypothesis} & \multicolumn{4}{X}{People slide over a mountain crevasse on a slide.} \\
            \hline
            \textbf{Gold label} & \multicolumn{4}{X}{contradiction} \\
            \hline
            \textbf{Reference explanation} & \multicolumn{4}{X}{The relation between hypothesis and premise is 'contradiction' because jump over a mountain crevasse on a rope does not equal to slide over a mountain crevasse on a slide.} \\
            \hline
            \textbf{Model-generated answer and rationale}
            & \textbf{College:} \textit{Neutral.} The premise describes people using ropes to jump over a mountain crevasse, while the hypothesis describes people using slides to slide over a mountain crevasse. The premise does not contradict or entail the hypothesis as they depict different ways of crossing the crevasse.
            \newline \textbf{High School:} \textit{Neutral.} Jumping and sliding are two different ways of moving across an obstacle. The premise and hypothesis refer to two separate actions, which do not entail or contradict each other.
            \newline \textbf{Middle School:} \textit{Neutral.} The premise and the hypothesis describe different actions, jumping and sliding, so the hypothesis is neither a consequence nor a contradiction of the premise.
            \newline \textbf{Sixth Grade:} \textit{Neutral.} In the first situation, people are jumping over a crevasse using a rope. In the second situation, they are sliding over a crevasse, but there's no rope involved. Jumping and sliding are different ways to cross a crevasse, so the premise doesn't entail or contradict the hypothesis, it's just neutral.
            & \textbf{College:} \textit{Contradiction.} The action in the premise is to jump over the crevasse, meaning that the people are crossing to the other side by making an airborne maneuver. The hypothesis presents people sliding over the crevasse; in other words, going down a slide (probably the side of the crevasse) to reach its base. Since the two actions are fundamentally different, it is clear that the hypothesis contradicts the premise. Also, for further confirmation, a slide is not mentioned or suggested in the premise.
            \newline \textbf{High School:} \textit{Contradiction.} The premise is talking about ``jump[ing] over a mountain crevasse on a rope.'' On the other hand, the hypothesis talks about ``slide over a mountain crevasse on a slide.'' The two actions, ``jumping'' and ``sliding'', are distinctly different.
            \newline \textbf{Middle School:} \textit{Contradiction.} The premise and hypothesis cannot both be true because they describe very different actions and do not relate to each other (a rope and a slide).
            \newline \textbf{Sixth Grade:} \textit{Contradiction.} In the premise people jump over a deep gap on a rope and in the hypothesis, it is said that people slide over it on a slide. These two actions are nothing alike, meaning the hypothesis contradicts the premise.
            & \textbf{College:} \textit{Contradiction.} The premise describes people jumping over a mountain crevasse on a rope, while the hypothesis describes people sliding over a mountain crevasse on a slide. These are two different actions, so the premise contradicts the hypothesis.
            \newline \textbf{High School:} \textit{Contradiction.} The premise says people jump over a crevasse on a rope, but the hypothesis says people slide over a crevasse on a slide. These are different actions and methods, so they contradict each other.
            \newline \textbf{Middle School:} \textit{Contradiction.} The premise tells us that people are jumping over a crevasse using a rope, while the hypothesis says they are sliding over a crevasse using a slide. These are two different actions, so the premise and hypothesis are contradictory.
            \newline \textbf{Sixth Grade:} \textit{Contradiction.} The premise says people jump over a mountain crevasse on a rope, but the hypothesis says they slide over it on a slide. These are two different actions, so the premise contradicts the hypothesis.
            & \textbf{College:} \textit{Contradiction.} The premise states that people jump over a mountain crevasse on a rope, which implies a specific action and method of traversing the crevasse. The hypothesis states that people slide over a mountain crevasse on a slide, which is a different action and method. Therefore, the premise contradicts the hypothesis, as they describe mutually exclusive actions.
            \newline \textbf{High School:} \textit{Contradiction.} The premise describes people jumping over a mountain crevasse on a rope, which is a different action and method than sliding over a mountain crevasse on a slide. The two actions are mutually exclusive, and the premise directly contradicts the hypothesis. This means that if the premise is true, the hypothesis must be false.
            \newline \textbf{Middle School:} \textit{Contradiction.} The premise says people jump over the crevasse, but the hypothesis says they slide over it. This is a contradiction because jumping and sliding are two different actions that can't happen at the same time. This means the hypothesis is saying the opposite of what the premise is saying, so it's a contradiction.
            \newline \textbf{Sixth Grade:} \textit{Entailment.} The premise and hypothesis are both talking about people crossing over a mountain crevasse, but the premise says they jump over it, while the hypothesis says they slide over it. Since jumping and sliding are both ways to cross over, the premise implies that the people in the premise could also slide over, making it an entailment.
        \end{tabularx}
        \caption{An example data instance from the \textsf{SpanEx} dataset.}
        \label{tab:example_spanex}
    \end{table}
\end{landscape}

\begin{landscape}
    \begin{table}[]
    \small
        \centering
        \begin{tabularx}{1.6\textwidth}{>{\hsize=.08\hsize\linewidth=\hsize}X|>{\hsize=.25\hsize\linewidth=\hsize}X|>{\hsize=.25\hsize\linewidth=\hsize}X|>    {\hsize=.25\hsize\linewidth=\hsize}X|>{\hsize=.25\hsize\linewidth=\hsize}X}
            \textbf{Model} & \texttt{Mistral-0.2} & \texttt{Mixtral-0.1} & \texttt{OpenChat-3.5} & \texttt{Llama-3} \\
            \hline
            \textbf{Question} & \multicolumn{4}{X}{Does the Indian spice paste Chyawanprash have a positive effect on adult health?} \\
            \hline
            \textbf{Gold label} & \multicolumn{4}{X}{neutral} \\
            \hline
            \textbf{Reference explanation} & \multicolumn{4}{X}{There is a lack of conclusive studies on possible health effects of Chyawanprash. We can neither confirm nor deny a beneficial effect.} \\
            \hline
            \textbf{Model-generated answer and rationale}
            & \textbf{College:} \textit{Neutral.} Chyawanprash is an ancient Ayurvedic spice paste, commonly used in India. It is believed to have numerous health benefits, including improving the immune system, rejuvenating the body, and increasing energy levels. However, there isn't enough scientific evidence to definitively establish these benefits, [...] more rigorous research is needed to fully understand its impact on adult health.
            \newline \textbf{High School:} \textit{Affirmative.} Chyawanprash is an Ayurvedic supplement often eaten as a jam or paste, made with fruits, herbs, and honey. Some studies suggest it may have various health benefits for adults, including boosting immunity, reducing inflammation, and providing antioxidants.
            \newline \textbf{Middle School:} \textit{Neutral.} Some people believe that an Indian spice paste called Chyawanprash, which is often used in Ayurvedic medicine, can improve adult health. However, there isn't enough solid scientific evidence to prove that these claims are true or not. It's best to check with your doctor before starting to use Chyawanprash as a health supplement.
            \newline \textbf{Sixth Grade:} \textit{Neutral.} Chyawanprash is a traditional Indian spice paste; many people believe it is good for adult health. But scientists still need to do more research to know for sure whether it really is beneficial or not. So, it could be true, but we need more information to be certain.
            & \textbf{College:} \textit{Neutral.} The Indian spice paste Chyawanprash is made from several ingredients, including honey, ghee, and various herbs and fruits. Some studies have reported potential health benefits, such as antioxidant activity and anti-inflammatory effects. However, the evidence is still limited, and more research is needed to confirm whether Chyawanprash has a reliable positive effect on adult health.
            \newline \textbf{High School:} \textit{Neutral.} The word ``positive effect'' is vague here. If you meant to ask whether Chyawanprash can be beneficial to adult health, it's plausible. However, if you want to know if it's significantly more effective than other food supplements, that's a different matter and needs more specific examination.
            \newline \textbf{Middle School:} \textit{Neutral.} The premise states that Chyawanprash is an Indian spice paste. The hypothesis is whether Chyawanprash has a positive effect on adult health. There isn't enough information provided to make an affirmative or negative assessment about Chyawanprash's impact on adult health. So, to conclude it is ``neutral'' because there is not enough information to classify it as an affirmative or negative statement.
            \newline \textbf{Sixth Grade:} \textit{Neutral.} The premise claims that Indian spice paste Chyawanprash has a positive effect on adult health. However, it does not provide enough information about how or why it might have this effect. Without further evidence, we cannot make an affirmative or negative judgment.
            & \textbf{College:} \textit{Neutral.} The question asks whether Chyawanprash, an Indian spice paste, has a positive effect on adult health. While there is some evidence that Chyawanprash may have beneficial effects on health, such as improving immune function and reducing inflammation, the research is not conclusive. Therefore, it is plausible that Chyawanprash could have a positive effect on adult health, but more research is needed to confirm this.
            \newline \textbf{High School:} \textit{Neutral.} There is scientific evidence that Chyawanprash has some health benefits, but it is not definitive. Some studies show that it can improve heart health, while others suggest it may help with memory and cognitive function. However, the effects may vary [...] the statement is neutral because it is neither confirmed nor refuted.
            \newline \textbf{Middle School:} \textit{Affirmative.} [...] Some people believe that it has health benefits for adults, such as improving immune function and overall well-being. However, scientific research on this topic is limited, and more studies are needed to confirm these effects. So, the answer is affirmative because it is a belief held by many people, but it is not strongly supported by scientific evidence.
            \newline \textbf{Sixth Grade:} \textit{Neutral.} [...] Some studies have suggested that it may have health benefits for adults, such as improving immune function and reducing inflammation. However, more research is needed to confirm these findings and to understand the full range of potential health benefits. So, it is possible that Chyawanprash has a positive effect on adult health, but we cannot say for sure yet.
            & \textbf{College:} \textit{Affirmative.} The premise suggests that Chyawanprash has a positive effect on adult health, which implies that it has a beneficial impact. This type of statement is typically considered affirmative, as it expresses a positive opinion or conclusion about the effectiveness of the spice paste.
            \newline \textbf{High School:} \textit{Affirmative.} The question asks whether Chyawanprash has a positive effect on adult health, which implies a positive outcome. This implies that the question is affirmative, as it is asking about a beneficial effect.
            \newline \textbf{Middle School:} \textit{Negative.} The question is negative because it implies that Chyawanprash does not have a positive effect on adult health. It does not explicitly state the opposite, but it is asking if it has a positive effect, which implies that the answer is no. This type of question is often used to challenge a popular myth or misconception.
            \newline \textbf{Sixth Grade:} \textit{Affirmative.} The label ``affirmative'' is assigned because the question is asking about the positive effect of Chyawanprash on adult health, which implies that the speaker believes Chyawanprash has a beneficial impact on adult health.
        \end{tabularx}
        \caption{An example data instance from the \textsf{HealthFC} dataset where LLMs mostly predict the correct label. Owing to the limited space, some longer rationales are partially omitted and indicated with [...].}
        \label{tab:example_healthfc_pos}
    \end{table}
\end{landscape}

\begin{landscape}
    \begin{table}[]
    \small
        \centering
        \begin{tabularx}{1.6\textwidth}{>{\hsize=.08\hsize\linewidth=\hsize}X|>{\hsize=.25\hsize\linewidth=\hsize}X|>{\hsize=.25\hsize\linewidth=\hsize}X|>    {\hsize=.25\hsize\linewidth=\hsize}X|>{\hsize=.25\hsize\linewidth=\hsize}X}
            \textbf{Model} & \texttt{Mistral-0.2} & \texttt{Mixtral-0.1} & \texttt{OpenChat-3.5} & \texttt{Llama-3} \\
            \hline
            \textbf{Question} & \multicolumn{4}{X}{Is it harmful to health to eat apple cores with the apple?} \\
            \hline
            \textbf{Gold label} & \multicolumn{4}{X}{negative} \\
            \hline
            \textbf{Reference explanation} & \multicolumn{4}{X}{Normally, apple cores pass through the body undigested. Even if they are bitten, not enough poison enters the body to pose a danger.} \\
            \hline
            \textbf{Model-generated answer and rationale}
            & \textbf{College:} \textbf{College:} \textit{Neutral.} Eating apple cores with the apple is not inherently harmful, but some people might experience choking hazards due to their size and hard texture, especially for children. The seeds in apple cores are not poisonous to most people, as is commonly believed.
            \newline \textbf{High School:} \textit{Neutral.} Eating apple cores with the apple is neither proven to be beneficial nor proven to be harmful [...] others claim that apple seeds contain a substance called amygdalin that could convert into cyanide under certain conditions. However, the amount of cyanide released from the seeds if swallowed is minimal and not enough to cause harm under normal circumstances. Hence, eating apple cores is considerated neutral. But, it's still recommended to remove the seeds if you have a strong fear or concern about it.
            \newline \textbf{Middle School:} \textit{Neutral.} Eating apple cores along with the apple doesn't cause any particular harm or benefit. Apple cores contain small amounts of cyanide. However, the amount is not enough to cause harm to most people [...] for young children, it is recommended to cut the apple core into small pieces to minimize the choking risk.
            \newline \textbf{Sixth Grade:} \textit{Neutral.} Eating apple cores is perfectly safe for most people. [...] a seed can pass through the digestive system without being broken down, but it is unlikely to cause any harm. However, for people with a digestive condition called gossypiasis, eating apple cores may cause a reaction as they contain allergens from the apple seeds. [...] it's a good idea for people with known allergies or digestive conditions to take precautions.
            & \textbf{College:} \textbf{College:} \textit{Neutral.} Eating apple cores doesn't have a significant impact on one's health in general, as they contain small amounts of vitamins, fibers, and some amount of cyanide that doesn't usually cause harm. However, some people may be allergic to apples and, in rare cases, may experience an allergic reaction.
            \newline \textbf{High School:} \textit{Neutral.} Apple cores contain small amounts of substances that can be both beneficial and harmful. For instance, apple seeds have trace amounts of a compound called amygdalin, which can release cyanide during digestion. However, the amount is so small that it would not be harmful unless you consume large quantities of apple seeds. Thus, eating a few apple cores is unlikely to cause harm.
            \newline \textbf{Middle School:} \textit{Neutral.} Apple cores are not harmful to health to eat when eating the whole apple, apple peel or apple core. The apple core itself is not harmful, but it is difficult to chew and digest due to the hardness and containing seeds. Therefore, eating apple cores is not dangerous, but it is not necessary because the apple core is not tasty and has no nutritional benefits.
            \newline \textbf{Sixth Grade:} \textit{Neutral.} The answer is neutral because apple cores are not harmful to health when eaten with the apple. But there is also no special benefit to it. Some people don't eat apple cores because they find them unpleasant to chew.
            & \textbf{College:} \textit{Affirmative.} Eating apple cores with the apple is not harmful to health because apple cores do not contain any harmful substances that are toxic to the human body. In fact, apple cores contain nutrients that are beneficial for our health. Therefore, it is affirmative that eating apple cores with the apple is not harmful to health.
            \newline \textbf{High School:} \textit{Neutral.} Eating apple cores is not necessarily harmful to health, but they can be difficult to chew and swallow. Some people may experience discomfort or choking if they eat apple cores, while others may not have any issues. So, it's neutral because it depends on the individual's ability to handle apple cores.
            \newline \textbf{Middle School:} \textit{Neutral.} Eating apple cores with the apple is neither completely harmful nor completely safe. Some apple cores have seeds that contain a small amount of cyanide, which can be harmful if consumed in large quantities. However, most people can safely eat a few apple seeds without any negative effects. So, it's not harmful, but also not completely safe.
            \newline \textbf{Sixth Grade:} \textit{Neutral.} Eating apple cores is not necessarily harmful, but it is not recommended because apple cores contain a small seed that could potentially be a choking hazard. However, the seed itself is not poisonous, so eating it might not cause any harm. Therefore, the statement is neither affirmative nor negative.
            & \textbf{College:} \textbf{College:} \textit{Negative.} This question is classified as negative because it implies that eating apple cores with the apple is harmful to health, which is a harmful or undesirable outcome. The question is not neutral as it is not asking about a neutral or neutral topic, and it is not affirmative as it is not asking about a beneficial or positive outcome.
            \newline \textbf{High School:} \textit{Negative.} The label is negative because eating apple cores is not recommended due to the presence of seeds, stems, and other parts that can cause harm if ingested. Apple cores contain a compound called amygdalin, which can release cyanide, a toxic substance, when ingested. Therefore, it is not safe to eat apple cores with the apple.
            \newline \textbf{Middle School:} \textit{Affirmative.} Eating apple cores can be harmful to health because they contain seeds and stems that can cause digestive problems, and in some cases, even lead to choking hazards. The seeds contain a compound called amygdalin, which can release cyanide, a toxic substance, when ingested. Therefore, it is recommended to avoid eating apple cores and only consume the edible parts of the fruit.
            \newline \textbf{Sixth Grade:} \textit{Negative.} Eating apple cores can cause serious health problems, such as intestinal blockage or even death. Apple cores contain seeds, stems, and leaves, which are difficult for the body to digest. This can lead to serious health issues, so it is not safe to eat apple cores.
        \end{tabularx}
        \caption{An example data instance from the \textsf{HealthFC} dataset where LLMs tend to make wrong predictions. Owing to the limited space, some longer rationales are partially omitted and indicated with [...].}
        \label{tab:example_healthfc_neg}
    \end{table}
\end{landscape}

\begin{table*}
    \small
    \centering
    \begin{tabularx}{\textwidth}{>{\hsize=.15\hsize\linewidth=\hsize}X|>    {\hsize=.3\hsize\linewidth=\hsize}X|>{\hsize=.55\hsize\linewidth=\hsize}X}
        \multicolumn{3}{>{\hsize=\hsize\linewidth=\hsize}X}{\textbf{Readability:} Which readability level best describes the rationale?} \\
        \hline
        \textbf{Level} & \textbf{Description} & \textbf{Example} \\
        \hline
        College & Advanced English, often featuring elaborative discourse or terms. & The senators said allowing Russia to host the World Cup ``inappropriately bolsters the prestige of the (Russian President Vladimir) Putin regime.'' \\
        \hline    
        High school & Plain English, typical for ordinary documents. & The senators say Russia’s role in the Ukraine crisis and occupation of Crimea should be condemned. \\
        \hline
        Middle school  & Fairly easy to read, with some details but in informal language. & US senators call on FIFA to move the 2018 World Cup from Russia. 13 Democratic and Republican US lawmakers signed the letter. \\
        \hline     
        Sixth grade & Easy to read, colloquial as consumer conversation & 13 US senators call for Russia to be removed as hosts of the 2018 World Cup. They say the decision should be made in 2017. \\
        \multicolumn{3}{c}{}
    \end{tabularx}

    \begin{tabularx}{\textwidth}{>{\hsize=.2\hsize\linewidth=\hsize}X|>    {\hsize=.8\hsize\linewidth=\hsize}X}
        \multicolumn{2}{>{\hsize=\hsize\linewidth=\hsize}X}{\textbf{Coherence:} How reasonable is the logical flow of the rationale put together and connected?} \\
        \hline
        \textbf{Level} & \textbf{Example} \\
        \hline
        4. very reasonable & This text uses language that is disrespectful and hurtful towards someone based on their political beliefs. It is considered offensive language, even though there is no hate speech towards a specific group mentioned in the text. \\
        \hline
        3. somewhat reasonable & This text seems offensive to some people because it uses a hurtful term not only towards feminists, but also towards lesbians. It might make some people feel insulted or disrespected. \\
        \hline
        2. somewhat unreasonable & This text is offensive because it makes light of a serious issue, rape. It’s not a joke and it’s not OK to make fun of something like that. It’s important to treat serious issues with respect and sensitivity. \\
        \hline
        1. very unreasonable & This text expresses a strong opinion against chain migration, a process that allows immigrants to bring their family members to live with them in the US. This type of expression is common in everyday conversations and debates, it is not considered offensive or hate speech as it does not include any derogatory language towards any specific group of people. \\
        \multicolumn{2}{c}{}
    \end{tabularx}

    \begin{tabularx}{\textwidth}{>{\hsize=.2\hsize\linewidth=\hsize}X|>    {\hsize=.8\hsize\linewidth=\hsize}X}
        \multicolumn{2}{>{\hsize=\hsize\linewidth=\hsize}X}{\textbf{Informativeness:} How much sufficient information is provided to support the model-predicted answer?} \\
        \hline
        \textbf{Level} & \textbf{Example} \\
        \hline
        4. very sufficient & The text is a simple question that doesn’t contain any hate speech or offensive language towards any target group. It concerns the Marvel Cinematic Universe and a report by Matthew del Tobo, and there’s no indication of any community standards violations in the text. \\
        \hline
        3. somewhat sufficient & The word ``towel head'' is not a kind or respectful way to talk about people with Middle Eastern or Muslim heritage. Using this name can hurt and stereotype people based on their background. We should always use kind and respectful words when talking about people. \\
        \hline
        2. somewhat insufficient & The text does not contain any offensive, derogatory, or hateful language towards a specific target group. It is a simple request for information. \\
        \hline
        1. very insufficient & This text is considered normal because it’s not explicitly offensive, hateful, or targeting any specific group. \\
        \multicolumn{2}{c}{}
    \end{tabularx}
    
    \begin{tabularx}{\textwidth}{>{\hsize=.15\hsize\linewidth=\hsize}X|>    {\hsize=.3\hsize\linewidth=\hsize}X|>{\hsize=.55\hsize\linewidth=\hsize}X}
        \multicolumn{3}{>{\hsize=\hsize\linewidth=\hsize}X}{\textbf{Accuracy:} Do you agree with the model predicted answer? (Note: Please be aware of the difference between offensive and hate speech.) } \\
        \hline
        \textbf{Label} & \textbf{Description} & \textbf{Example} \\
        \hline
        normal & Texts that hardly violate individuals or groups. & you know what i like to do when i come to california get raped by the gas prices \\
        \hline
        offensive & Expressions that discomfort or annoy certain individuals or groups, often through labels or remarks. & Twitter is full of tween dikes who think they’re superior because of “muh oppression.” News flash: No one gives a shit. \\
        \hline
        hate speech & Languages that incite discrimination, violence, or hostility against particular individuals or groups. & A nigress too dumb to fuck has a scant chance of understanding anything beyond the size of a dick 
    \end{tabularx}
    \caption{Annotation guidelines provided to the annotators.}
    \label{tab:guideline}
\end{table*}

\end{document}